\documentclass{article}

\PassOptionsToPackage{numbers,sort&compress}{natbib}

    \usepackage[preprint]{neurips_2025}

\usepackage[utf8]{inputenc} 
\usepackage[T1]{fontenc}    
\usepackage[colorlinks]{hyperref}       
\usepackage{url}            
\usepackage{booktabs}       
\usepackage{amsfonts}       
\usepackage{wrapfig}        
\usepackage{nicefrac}       
\usepackage{microtype}      
\usepackage{xcolor}         
\definecolor{iccvblue}{rgb}{0.21,0.49,0.74}
\usepackage{caption}
\usepackage{algorithm}
\usepackage{algpseudocode}
\usepackage{enumitem}
\usepackage{tikz}
\usetikzlibrary{shapes, arrows}
\usepackage{tikz-3dplot}
\usetikzlibrary{positioning, shapes.geometric}
\usepackage{amsmath}
\usepackage{tcolorbox}
\usepackage{lipsum} 
\usepackage[]{mdframed}
\usepackage{cleveref}
\usepackage{subcaption}
\usepackage{multirow}
\definecolor{pastelgreen}{rgb}{0.75, 0.95, 0.75}
\definecolor{pastelblue}{rgb}{0.75, 0.85, 0.95}
\PassOptionsToPackage{table}{xcolor}

\newcommand{\formattedparagraph}[1]{\noindent \textbf{#1}}
\tcbuselibrary{breakable}

\title{Interactive Video Generation via Domain Adaptation}

\author{%
    Ishaan Rawal\textsuperscript{1\,2\,}\thanks{Work done as a graduate research assistant in the Visual and Spatial AI Lab, College of PVFA} \quad
    Suryansh Kumar\textsuperscript{1\,2\,3} 
    \\
    \textsuperscript{1}\small{College of PVFA} \quad
    \textsuperscript{2}\small{Dept.~of Computer Science \& Engineering} \quad
    \textsuperscript{3}\small{Dept.~of Electrical \& Computer Engineering} \\
    \small{Texas A\&M University} \\
  \small{\texttt{\{ishaanrawal, suryanshkumar\}@tamu.edu}} \\
}

\definecolor{amber}{rgb}{1.0, 0.68, 0.0}

\begin{document}

\maketitle

\begin{abstract}

Text-conditioned diffusion models have emerged as powerful tools for high-quality video generation. However, enabling Interactive Video Generation (IVG), where users control motion elements such as object trajectory, remains challenging. Recent training-free approaches introduce attention masking to guide trajectory, but this often degrades perceptual quality. We identify two key failure modes in these methods, both of which we interpret as domain shift problems, and propose solutions inspired by domain adaptation. First, we attribute the perceptual degradation to \emph{internal covariate shift} induced by attention masking, as pretrained models are not trained to handle masked attention. To address this, we propose mask normalization, a pre-normalization layer designed to mitigate this shift via distribution matching. Second, we address \emph{initialization gap}, where the randomly sampled initial noise does not align with IVG conditioning, by introducing a temporal intrinsic diffusion prior that enforces spatio-temporal consistency at each denoising step. Extensive qualitative and quantitative evaluations demonstrate that mask normalization and temporal intrinsic denoising improve both perceptual quality and trajectory control over the existing state-of-the-art IVG techniques.

\end{abstract}

\section{Introduction}\label{sec:intro}


Controlling text-to-video diffusion models has enabled new forms of human-AI interaction for creative and simulation-based tasks. A key challenge in this area is trajectory control, where users specify the subject's spatio-temporal position movement across frames, commonly referred to as Interactive Video Generation (IVG)~\cite{peekaboo, trailblazer}. At present, optimal trajectory control in IVG remains a challenging problem. A suitable solution to IVG must maintain approximate object's size and position across frames, ensuring spatial consistency despite appearance changes, and generate smooth transitions, even under complex dynamic motion.

\begin{figure}[t]
\centering
\includegraphics[width=\textwidth]{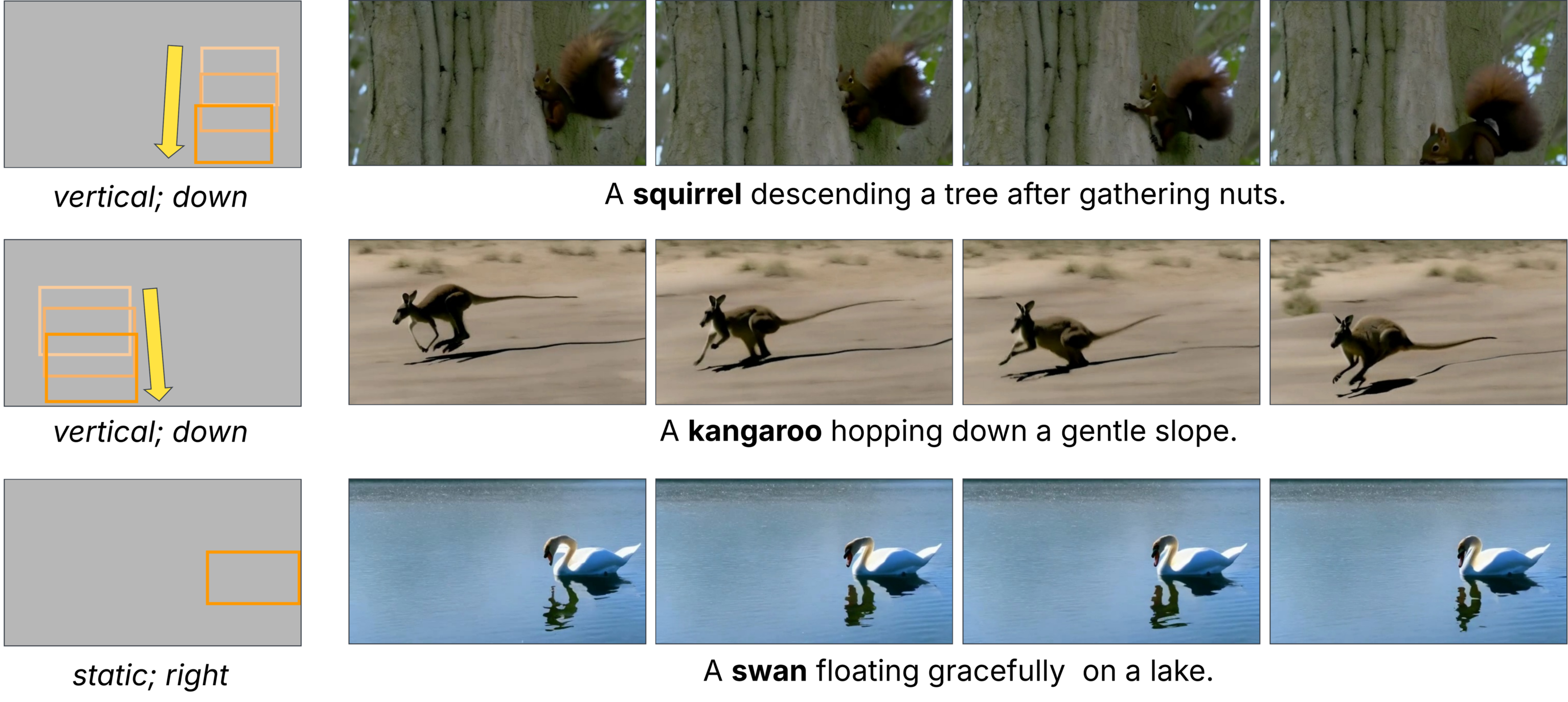}
\caption{\small \label{fig:intro-fig} The task is to generate videos conditioned on a text prompt and a bounding box trajectory mask. Each row shows one instance: the gray panel on the left visualizes the bounding box trajectory (darker boxes appear later in time; arrows indicate direction for clarity), and the following frames show the generated video. The text prompt is shown below each video, with the subject in bold. Our model generates coherent, trajectory- and caption-aligned videos—without any training or finetuning.}
\end{figure}

\noindent
Many popular methods for IVG require retraining or fine-tuning~\cite{jeong2024vmc, chen2023motion, wu2023tune, wu2024lamp, yang2024direct}, often incorporating additional components such as Low-Rank Adaptation (LoRA)~\cite{hu2021lora, zhao2024motiondirector, wu2024customcrafter, ren2024customize} and learned embeddings~\cite{li2024animate, li2025director3d, liu2024video, niu2024mofa, wangboximator, zhang2023adding, zhang2024tora, wang2024motionctrl, wei2024dreamvideo}. These approaches are often prohibitive, as they rely on custom-annotated datasets, incur significant computational costs, and typically require a reference image, limiting their applicability. To counter these limitations, training-free solutions have been proposed recently~\cite{peekaboo, jeong2024dreammotion, yan2021videogpt, trailblazer, wu2024freeinit, freetraj}. These approaches take per-frame bounding boxes as input and benefit from the rapid progress in video generation models, as shown in \cref{fig:intro-fig}. They achieve trajectory control through customized noise initialization techniques~\cite{qiufreenoise, freetraj, wu2024freeinit} and attention masking~\cite{peekaboo, jeong2024dreammotion, freeinit, xiao2024video, trailblazer}. Despite these advancements, a critical weakness persists: these methods often trade off trajectory control for perceptual quality. This tradeoff limits their applicability, motivating the need for new techniques that improve video generation quality while preserving effective trajectory control.


In this paper, we introduce a principled approach to enhance training-free IVG using attention masking, while explicitly enforcing spatio-temporal coherence across frames. We begin by identifying two key limitations in existing methods. First, we find that video diffusion models are not trained with masked attention; applying attention masks during inference introduces \textbf{internal covariate shift} which was overlooked in prior works. Second, it is known that the imperfect nature of the diffusion noising process causes information leakage in the noised latent, known as the \textbf{initialization gap}~\cite{freeinit}. As a result, the model implicitly expects trajectory-conditioned latent inputs at inference, yet most, if not all, IVG methods do not account for this discrepancy. Having identified domain shifts unique to masked video diffusion models, we ask:

\begin{tcolorbox}[boxrule=1pt, sharp corners, parbox=false, breakable, colback=gray!5]
\textit{Can we interpret and ultimately improve both control and perceptual quality in masked video diffusion models through domain adaptation at inference time?}
\end{tcolorbox}




Let us formalize this idea. Assume a source domain distribution as $\mathcal{D}_s$ and a target domain distribution as $\mathcal{D}_t$. As alluded to above, the challenges with current IVG methods involve two independent domain shift problems: \textbf{(i)} internal covariate shift between unmasked attention outputs ($\mathcal{D}_s$) and masked attention outputs ($\mathcal{D}_t$), and \textbf{(ii)} a distribution gap between randomly initialized latents ($\mathcal{D}_s$) and IVG-conditioned latents ($\mathcal{D}_t$). In both cases, the objective is to minimize the discrepancy between $\mathcal{D}_s$ and $\mathcal{D}_t$. We adopt a two-pronged approach to address these problems independently.

First, to resolve the internal covariate shift induced by masking, we propose \textbf{mask normalization}, a novel pre-normalization layer. Inspired by style transfer, a classic example of domain adaptation, mask normalization aligns the feature distributions between the outputs of unmasked attention (as the \textit{content image}; $\mathcal{D}_s$) and masked attention (as the \textit{reference image}; $\mathcal{D}_t$). Our analysis shows that mask normalization effectively reduces the variance shift in activations caused by attention masking.

Second, to address the initialization gap, we draw inspiration from Deep Image Prior (DIP)~\cite{deepimageprior}, a well-known domain adaptation method in image denoising. At each denoising step, we treat the current latent as a \textit{noisy image} ($\mathcal{D}_s$) and the expected IVG-conditioned latent as the \textit{clean image} ($\mathcal{D}_t$). Similar to DIP, we leverage the \textbf{intrinsic diffusion prior} to iteratively denoise the latent. This process is further guided by a correlative \textbf{temporal prior} using classifier-free guidance, which explicitly enforces spatio-temporal coherence across tokens and frames.

In summary, this paper presents a new perspective on IVG through the lens of domain adaptation. We introduce the following contributions to improve training-free IVG with masked diffusion models: \textbf{(1)} {Mask normalization}, a novel pre-normalization layer that mitigates internal covariate shift by aligning feature distributions; \textbf{(2)} {Intrinsic denoising}, a modification to the diffusion sampling process that leverages the denoising prior; and \textbf{(3)} A {classifier-conditioned temporal prior} that guides intrinsic denoising to ensure spatio-temporal coherence. Our approach achieves state-of-the-art performance, outperforming all prior training-free IVG methods.







\section{Related Works}\label{sec:related_works}



In recent years, diffusion models have surpass GANs~\cite{goodfellow2014generative} in data generation quality, particularly in image and video synthesis. Unlike GANs, which often suffer from mode collapse and training instability, diffusion models benefit from a well-defined denoising process that enables fine-grained control and high-fidelity visual generation. Given our paper’s focus on interactive video generation (IVG) using diffusion models, we restrict our discussion to directly relevant approaches.


\formattedparagraph{Video Diffusion Models.}
Text-based video generation using latent diffusion models has advanced significantly in recent years~\cite{brooks2024video, chen2023control, ho2022video, zhou2022magicvideo, singer2022make}. While these models achieve high semantic alignment with textual prompts, they lack fine-grained spatiotemporal control at the frame level. More recent approaches have attempted to address such a limitation by incorporating additional guidance signals, such as depth maps~\cite{esser2023structure}, target motion trajectories~\cite{chen2023motion, hu2023lamd}, or a combination of these modalities~\cite{wang2023videocomposer}. Yet, these methods typically require either re-training the base diffusion model or integrating an external adapter aligned with grounded spatiotemporal data---both of which are computationally intensive and costly.

In contrast, zero-shot video generation approaches eliminate the need for re-training. For instance, Text2Video-Zero~\cite{khachatryan2023text2video} integrates optical flow guidance into a pre-trained image diffusion model to improve temporal consistency. ControlVideo~\cite{zhang2023controlvideo} extends control mechanisms by supervising motion through a sequence of structured frames (e.g., depth maps, stick figures). Free-Bloom~\cite{huang2023free} employs a large language model (LLM) in conjunction with a text-to-image model to enhance coherence in generated videos. Nevertheless, these approaches rely on pre-trained image models fine-tuned on grounded data and are not readily compatible with off-the-shelf video-generation models.

\formattedparagraph{Guidance based Video Generation.} Recent studies in video generation via diffusion models have explored incorporating spatial and stylistic control into text-to-image generation. These approaches are either training-based~\cite{li2023gligen} or training-free based~\cite{agarwal2023star, cao2023masactrl, epstein2023diffusion, phung2024grounded, peekaboo, trailblazer, freetraj, direct-a-vid}. Training-based methods require substantial computational resources and rely on spatially grounded datasets to optimize the diffusion model for enhanced spatial control. In contrast, training-free approaches manipulate spatial and cross-attention maps through energy function-guided diffusion or masking techniques. 
Our approach aligns with the latter category of methods, particularly those such as~\cite{peekaboo, trailblazer, freetraj, direct-a-vid}, which employ large language models (LLMs) to generate bounding box trajectories for objects across video frames conditioned on textual descriptions. These methods integrate off-the-shelf video generation models with auxiliary mask-based guidance modules to facilitate localized control. Yet, the critical limitation remains: maintaining spatiotemporal consistency across frames. Despite their effectiveness in generating semantically relevant motion, these methods often exhibit temporal artifacts or inconsistencies in object appearance and motion continuity. This undermines the visual coherence essential for interactive video generation (IVG).



\section{Method}\label{sec:method}

In this section, we present our approach to improving IVG by adapting a pre-trained text-to-video diffusion model for joint text and trajectory control via domain adaptation. We begin by introducing masking-based trajectory control (\cref{sec:masking_review}) and analyze how masking leads to a variance shift in activations, degrading perceptual quality (\cref{sec:var_shift}). To address this, we propose mask normalization (\cref{sec:masknorm}). Next, we uncover the limitation of initialization gap that arises during inference in video diffusion models (\cref{sec:noise_refinement}) and introduce a lightweight temporal prior (\cref{sec:temporal_prior}) to bridge this gap through temporal intrinsic denoising (\cref{sec:diffusion_sampling}).

\formattedparagraph{Problem setup.} Our goal is to generate an $N$-frame video $z_0 \in \mathbb{R}^{N \times H \times W \times 3}$, given: \textbf{\textit{(i)}} a text prompt $y$ describing the scene and specifying a single subject $s$, i.e., the foreground, and \textbf{\textit{(ii)}} a bounding box mask $b \in \{0, 1\}^{N \times H \times W}$ indicating the subject's trajectory across frames. Here, $H, W$ symbolizes height and width of image frame. We assume $1$, $0$ to be foreground and background, respectively, where the words "foreground" and "subject" are used interchangeably. The output of the diffusion model at step $t$ is denoted by $z_t$, where $ t \in \{1, \ldots, T\}$. Furthermore, we assume that we have a pre-trained and frozen T2V diffusion model. In this work, we adopt Peekaboo's \cite{peekaboo} masking scheme, summarized in the following section.

\subsection{Masking-based trajectory control}
\label{sec:masking_review}
We begin by reviewing masking-based trajectory control, a method for enabling trajectory control in text-to-video models by introducing spatial and temporal attention masks, without requiring any finetuning. We use the masking scheme proposed in Peekaboo paper \cite{peekaboo}, where only foreground-to-foreground and background-to-background tokens can attend to each other. Specifically, we assume access to two binary masks: \( M_v \in \{0,1\}^{T \times l_t \times 1} \), representing the user-specified bounding box trajectory, and \( M_y \in \{0,1\}^{l_t \times 1} \), representing the conditioning prompt. In both masks, foreground tokens are assigned the value \( 1 \), and background tokens \( 0 \). 

The spatial self-attention \( M_{\text{self}}[i] \) and the spatial cross-attention  \( M_{\text{cross}}[i] \) at timestep \( i \) are defined as:
\[
M_{\text{self}}[i] = (M_v[i] \cdot M_v[i]^\top) + \big((1 - M_v[i]) \cdot (1 - M_v[i])^\top\big)
\]
\[
M_{\text{cross}}[i] = (M_v[i] \cdot M_y[i]^\top) + \big((1 - M_v[i]) \cdot (1 - M_y[i])^\top\big)
\]

The masks are downsampled to match the dimensions of the attention layers. Temporal self-attention masks are similarly constructed to spatial self-attention masks. These masks are then applied to the decoder attention layers of the U-Net \cite{ronneberger2015u} during video generation to enforce trajectory control. The masks are applied only for the first few denoising steps, referred to as ``frozen steps''.

\subsection{Masking-induced internal covariate shift}
\label{sec:var_shift}

Latent video diffusion models like Stable Video Diffusion and Zeroscope generate videos by progressively denoising a randomly initialized latent using convolutional and attention-based networks. Unlike language models trained with masked objectives, these models are not exposed to attention masking during training. However, recent trajectory control methods often introduce attention masks during inference. To understand the impact of such masking, we analyze the variance of attention outputs. Specifically, we extract activations from the transformer layer in the second decoder block of Zeroscope to observe how feature variance evolves across diffusion steps. We fix the number of frozen steps to four and compare naive masking with our method, which applies mask normalization and temporal intrinsic denoising (explained later) to reduce variance shifts introduced by masking.

\begin{figure}[h]
    \centering
    \begin{subfigure}[t]{0.41\textwidth}
        \includegraphics[width=\linewidth]{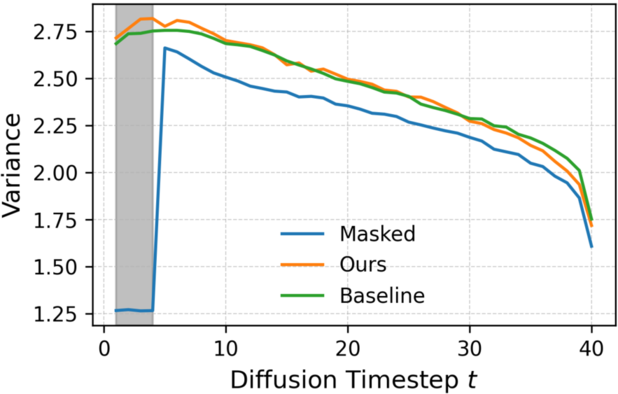}
        \caption{Variance distribution curve}
        \label{fig:var_vs_t}
    \end{subfigure}
    \hfill
    \begin{subfigure}[t]{0.41\textwidth}
        \includegraphics[width=\linewidth]{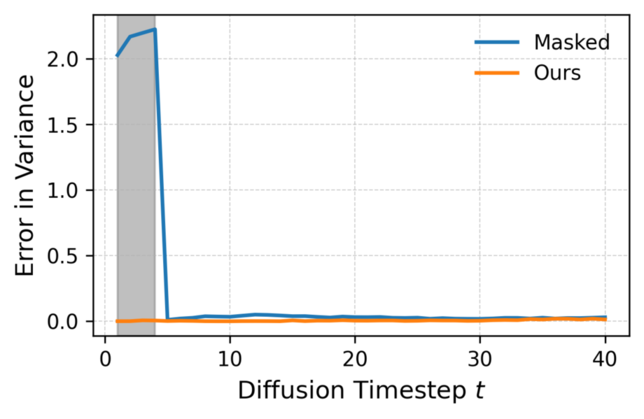}
        \caption{Variance error curve}
        \label{fig:mse_var_vs_t}
    \end{subfigure}
    \caption{\small \textbf{Effect of attention masking on activation variance.} The attention layers in the first 4 diffusion steps are masked for trajectory control (indicated in \textcolor{gray}{gray}). No masking is applied for the baseline configuration. 
(a) The masked variant (\textcolor{blue}{blue}) exhibits internal variance shift relative to the baseline (\textcolor{green}{green}), whereas our method (\textcolor{orange}{orange}) results in more aligned variance distribution, even in presence of masking.  
(b) The lower variance error (measured as the mean squared error with respect to the baseline) in our method, compared to the masked configuration, confirms that our method mitigates the variance shift due to attention masking.}
    \label{fig:variance_analysis}
\end{figure}

It can be observed from \cref{fig:var_vs_t} statistical analysis plot that applying mask in the early steps causes a notable drop in activation variance relative to the baseline (unmasked), indicating internal covariate shift. Interestingly, immediately after masking ends, the variance spikes sharply (at step 5), before decaying relatively smoothly. In contrast, our method maintains variance levels more consistent with the baseline throughout. Fig.~\ref{fig:mse_var_vs_t} quantifies this behavior in terms of mean squared error with respect to the baseline. The variance error increases progressively with masking, while our method consistently maintains low error, empirically supporting its ability to mitigate the variance shift introduced by attention masking. \emph{These results are surprising and interesting: despite the presence of layer normalization in transformer blocks, it does not appear sufficient to counteract the variance shift introduced by early attention masking.}

Why does early-step variance shift matter? Prior work by \citet{freeinit} shows that video diffusion models suffer from ``initialization gap'': the randomly sampled latents at inference time do not match the model's training distribution due to imperfections in the diffusion noising process, which harms generation quality. We extend this idea to activations, suggesting that early out-of-distribution (OOD) activations caused by masking-induced covariate shift lead to updates that remain ODD, compounding the drift and degrading the final output. One might consider deferring attention masking to later diffusion steps to avoid early variance shift, but this strategy is ineffective: the model commits to low-level structural features, such as object layout and motion trajectory, early in the generation process. The challenge is further compounded by the constraint that we aim to correct the variance shift without modifying network weights or altering the layer-wise feature distribution. This presents a core challenge—largely overlooked in prior work—for masked diffusion models: \textbf{How can we mitigate internal covariate shift without training?}

\subsubsection{Mask normalization}
\label{sec:masknorm}
Our analysis in the previous section shows that layer normalization alone is insufficient to correct the internal covariate shift induced by attention masking, which degrades generation quality. To address this, we propose \textbf{mask normalization}. It is a  lightweight pre-normalization layer that aligns the distribution of masked attention outputs with their unmasked counterparts. Mask normalization is therefore applied before the residual connection and final normalization layer in each masked attention block, with the goal of preserving expected variance and reducing covariate shift. This leads to improved generation quality in trajectory-conditioned video diffusion under masking.

This naturally invites a broader conceptual perspective on our design. To ground our method in an intuitive framework, we draw inspiration from the style transfer literature, where generation quality is improved by aligning feature statistics between domains. Analogously, in our case, the masked attention outputs serve as the \textit{reference distribution} defining the spatial location of the subject in time, while the unmasked attention outputs resemble the \textit{content distribution} to be adapted. This analogy provides a useful lens through which to interpret mask normalization as a form of distribution matching that regularizes attention dynamics under interactive constraints. Specifically, for the output of each masked attention layer \(A_m\), we compute the corresponding unmasked attention output, \(\mathcal{A}_u\). If \(Q\), \(K\), and \(V\) denote the query, key, and value, \(\mathcal{M}\) is the multiplicative binary mask, and \(\sigma\) is the softmax function, we define:
\begin{equation*}
\mathcal{A}_m = \sigma\left(\frac{QK^\top}{\sqrt{d}} \circ \mathcal{M}\right) V \quad \text{and} \quad \mathcal{A}_u = \sigma\left(\frac{QK^\top}{\sqrt{d}}\right) V 
\end{equation*}
\begin{wrapfigure}{Rh}{0.40\textwidth}
\centering
\vspace{-0.5em}
\includegraphics[width=0.40\textwidth]{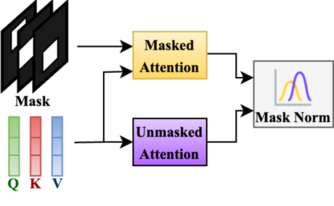}
\caption{\label{fig:masknorm}Mask Normalization aligns the empirical cumulative distributions of masked and unmasked attention outputs using EFDM (Q, K, V are query, key, value tensors of attention).}
\end{wrapfigure}

For simplicity, we assume that \(\mathcal{A}_m\) and \(\mathcal{A}_u\) are \(N \times 1\)-dimensional vectors. Exact Feature Distribution Matching (EFDM)~\cite{efdm}, based on Exact Histogram Matching~\cite{coltuc2006exact}, is a popular method in style transfer for matching the empirical Cumulative Distribution Function of two distributions. In EFDM, each element of \(\mathcal{A}_m\) is replaced by the value from \(\mathcal{A}_u\) that has the same rank (i.e., the \(k\)-th smallest element of \(\mathcal{A}_m\) is mapped to the \(k\)-th smallest element of \(\mathcal{A}_u\)).

Thus, EFDM essentially rearranges the values in \(\mathcal{A}_u\) so that their relative ordering matches that of \(\mathcal{A}_m\). This can also be viewed as a discrete 1D Monge optimal transport problem~\cite{monge1781memoire} with quadratic cost, where the normalized unmasked attention output is mapped to the normalized masked attention output, followed by unnormalization. Since rearrangement within a layer does not alter the distribution, Mask Normalization preserves the original layer-wise distribution while also reducing the average variance shift, as shown in \cref{fig:variance_analysis}. 

We illustrate mask normalization in \cref{fig:masknorm}, which is applied independently to each head in a batch along the last dimension. Since masking (and thus mask normalization) is applied only during the early diffusion steps, it introduces minimal overhead to the inference. Furthermore, the computation of EFDM using Sort Matching~\cite{rolland2000fast} is efficient, with a computational complexity of \(O(n \log n)\). Next, we go into the  limitation of initialization gap that arises in video diffusion models at inference time. 

\subsection{Noise refinement with diffusion prior}
\label{sec:noise_refinement}
The denoising process in diffusion models is strongly influenced by the initial noise because of imperfect noising schedules~\cite{lin2024common,freeinit}. Therefore, during inference, randomly initialized latent noise fails to capture the high inter-frame, low-frequency correlations inherent in videos. To bridge this initialization gap, current methods adopt techniques such as iterative frequency refinement of latent noise~\cite{freetraj} or inversion (i.e., noising) of a reference video to initialize the latent~\cite{motion-zero}. However, these approaches require significant manual intervention and do not leverage the intrinsic prior of the diffusion process---a rich and free supervision signal.

We propose a simple solution to bridge this initialization gap by drawing inspiration from the image denoising literature. Image denoising again can be viewed as a domain adaptation problem, i.e., mapping a noisy image to its clean counterpart. A popular approach in this space is the Deep Image Prior~\cite{deepimageprior, dreamclean}, which leverages the architecture of a neural network itself as a prior for generating realistic images. Motivated by these insights, we model the randomly initialized latent as the \textit{noisy image}, and the desired correlated latent at a given diffusion timestep as the \textit{clean image}. We refer to this process as \textbf{temporal intrinsic denoising} to distinguish it from the standard denoising process in diffusion models, which involves stochastic sampling (elaborated in \cref{sec:diffusion_sampling}). A key distinction is that, unlike the Deep Image Prior which does not incorporate any prior knowledge about the restoration process, our approach incorporates lightweight classifier guidance to steer intrinsic denoising using the text prompt and bounding box masks---described next in subsections that follows.

\subsubsection{Temporal prior}
\label{sec:temporal_prior}
We employ user-specified bounding box masks for trajectory control to guide the classifier during noise refinement via intrinsic denoising. We compute the Pearson correlation between foreground pixels across consecutive frames to evaluate temporal consistency. Since bounding box sizes may vary, we sample a fixed number of pixel values from the foreground regions of successive frames.

Let $u_t^i \in \mathbb{R}^{h^i \times w^i \times 3}$ be the region of size $h^i \times w^i$ extracted from the intermediate diffusion output $z_t$ corresponding to the bounding box $b[i]$ of the $i$\textsuperscript{th} frame. We define the temporal consistency metric $\tau$  as,
\begin{equation}
  \tau\big(\tilde{b} , z_t\big) = \frac{1}{N-1} \sum_{i=1}^{N-1} \rho\Big( c\big(u_t^i, h_{\min}^i, w_{\min}^i\big),\, c\big(u_t^{i+1}, h_{\min}^i, w_{\min}^i\big) \Big),
\label{eqn:temporal_prior}
\end{equation}
$\rho(\cdot,\cdot)$ denotes the Pearson correlation function, and
$h_{\min}^i = \min\{h^i,\, h^{i+1}\}$, $w_{\min}^i = \min\{w^i,\, w^{i+1}\}$. Here, the function $c(\cdot, h_{\min}^i, w_{\min}^i)$ uniformly samples $h_{\min}^i \times w_{\min}^i$ pixels. We use $\tilde b$ to denote that we derive explicit cross-token and cross-frame prior from $b$.

Existing masking-based trajectory control methods impose spatio-temporal control by masking foreground and background tokens within attention layers. However, U-Net-based diffusion models lack explicit attention mechanisms for \emph{cross-frame, cross-token} interactions. As a result, these methods are limited in their ability to jointly control spatial and temporal dynamics. In contrast, as demonstrated by our formulation of $\tau$, we impose supervision jointly across tokens and frames, thereby introducing a explicit spatio-temporal prior.

\subsubsection{Temporal Intrinsic Denoising (TID)}
\label{sec:diffusion_sampling}

In this section, we introduce intrinsic denoising (ID)---a modification to the diffusion inference process to leverage the intrinsic prior of diffusion and modify it to leverage the explicit temporal prior described in \cref{sec:temporal_prior} (temporal intrinsic denoising, or TID) to bridge the initialization gap. Intrinsic denoising was originally described as variance preserving sampling \cite{dreamclean} for image restoration. Before each diffusion sampling (i.e. denoising) step, intrinsic denoising performs the following step $M$ times:
\begin{equation}
z_{t}^{m} = z_{t}^{m-1} + 
\eta_l \nabla_z \log p_t(z_{t}^{m-1}) + \eta_k \epsilon_{k}^{m}, 
\end{equation}
where $\eta_l = \gamma \, (1 - \bar\alpha_t) $  and $\eta_k = \sqrt{\gamma \, (2-\gamma)}\, \sqrt{(1 - \bar\alpha_t)}$ for $0 < \gamma < 1$, and $m \in \{1, 2, \cdots, M\}$. However, it does not account for classifier-conditioned generation. 
Taking inspiration from classifier-guidance~\cite{dhariwal2021diffusion}, we modify the equation as:
$$z_{t}^{m} = z_{t}^{m-1} + 
\eta_l \nabla_z \log g_t\big(z_{t}^{m-1} \, |\, \tilde b \big) + \eta_k \epsilon_{k}^{m} $$

But we know that $z_t$ is conditioned on $y$ and $b$. Replacing this, we get, $$z_{t}^{m} = z_{t}^{m-1} + 
\eta_l \nabla_z \log g_t \left(z_{t}^{m-1} \, |\, y , b  \, \Big| \, \tilde b \right) + \eta_k \epsilon_{k}^{m}$$
Expanding using Bayes' rule, we arrive at the following,

\begin{equation*}
z_{t}^{m} = z_{t}^{m-1} + 
\eta_l \bigg[ \nabla_z  \log g_t \left( \tilde b \, \Big| \, z_{t}^{m-1} \, |\, y , b \right) + \nabla_z  \log p_t(z_{t}^{m-1} \, |\, y , b) \bigg] + \eta_k \epsilon_{k}^{m}    
\end{equation*}

We can use the temporal prior in \cref{eqn:temporal_prior} to approximate $\nabla_z  \log g_t \big( \tilde b \, \Big| \, z_{t}^{m-1} \, |\, y , b \big) \approx \nabla_z \tau\big(\tilde b, \, z_{t}^{m-1} \, |\, y , b \big)$. Further, rather than computing $\tau\big(\tilde b ,\cdot \big)$ as a function of $z_t^{m-1}$, we calculate it on one-shot clean approximation of the clean sample, $\hat{z}_{0,t}^{m-1}$~\cite{bansal2024universal}  using Tweedie's formula~\cite{efron2011tweedie}, given as : $\hat{z}_{0,t}^{m-1}= \frac{z_t^{m-1} - \sqrt{1-{\alpha}_t} \, \hat{\epsilon}_{\theta}(z_{t}^{m-1} \, |\, y, b)}{\sqrt{\alpha}_t}, $ where $\hat\epsilon_\theta = (1 + \omega) \, \epsilon_\theta(z_{t}^{m-1} \, |\, y, b)  - \omega \, \epsilon_\theta(z_{t}^{m-1})$, and $\epsilon_\theta$ is the noise predicted by the diffusion model, and $\omega$ is the guidance scale. Morover, we scale the classifier-guided gradients with $c_g$. Putting everything together, we get the final expression for temporal intrinsic denoising:
\begin{equation}
z_{t}^{m} = z_{t}^{m-1} + \eta_l \bigg[ 
\underbrace{c_g \nabla_z \tau \big( \tilde b ,\, \hat{z}_{0,t}^{m-1} \big)}_{\text{{classifier} guidance (temporal prior)}}  +  \;\;\;\;\ \underbrace{\nabla_z \log p_t \big(z_{t}^{m-1} \, |\, y , b \big)}_{\text{classifier-free guidance}}
\bigg] +   \eta_k \epsilon_{k}^{m}
\label{eqn:intrinsic_denoising}
\end{equation}
We use temporal intrinsic denoising to refine the latent noise before sampling for the next timestep, as illustrated in \cref{alg:pseudocode}. In practice, we apply temporal prior to the latent space, and not the pixel space, of the diffusion model for computational efficiency. Although seemingly similar, our approach is fundamentally different from recursive denoising~\cite{lugmayr2022repaint, gengmotion}, which repeats the diffusion sampling steps multiple times for each time step for stable optimization. Intrinsic denoising, on the other hand, is theoretically motivated to leverage the inductive prior of the diffusion model and process to guide the corrupted low-probability latents towards the nearby high-probability region~\cite{dreamclean} while being faithful to both classifier and classifier-free conditioning.


\begin{wrapfigure}{r}{0.5\textwidth}
\begin{minipage}{\linewidth}
\vspace{-4.75em}
\begin{algorithm}[H]
\small{
\caption{\label{alg:pseudocode} Diffusion Sampling with TID}
\begin{algorithmic}
\For{$t = T$ \textbf{downto} $1$}
    \State $z_t^0 \gets z_t$ 
    \For{$m = 1$ \textbf{to} $M$}
        \State $\quad$ \textbf{// TID (\cref{eqn:intrinsic_denoising})}
         \State $\quad$ $\eta_l \gets \gamma \, (1 - \bar\alpha_t) $
         \State $\quad$ $\eta_k \gets \sqrt{\gamma \, (2-\gamma)}\, \sqrt{(1 - \bar\alpha_t)}$
        \State $\quad$ $\mathcal{G}_g \gets \nabla_z \tau \Bigl(\tilde b , \, \hat{z}_{0,t}^{m-1}\Bigr)$
        \State $\quad$ $\mathcal{G}_p \gets \nabla_z \log p_t\bigl(z_t^{m-1} \, |\, y, b\bigr)$ 
        \State $\quad$ $\epsilon_k^m \gets \text{sample\_noise(\,)}$ 
        \State $\quad$ $z_t^m \gets z_t^{m-1} + \eta_l \Bigl[ c_g \, \mathcal{G}_g + \mathcal{G}_p \Bigr] + \eta_k \epsilon_k^m$ 
    \EndFor 
    \State $z_t \gets z_t^{M}$
    \State \textbf{// Diffusion Sampling}
    \State $z_{t-1} \gets \textrm{DDIMStep}(z_t, y, b, t)$ 
\EndFor
\State \textbf{Output:} $z_0$
\end{algorithmic}
}
\end{algorithm}
\end{minipage}
\end{wrapfigure}

\section{Experiment and Evaluations}
\label{sec:experiments}
    In this section, we present the main qualitative and quantitative results demonstrating the effectiveness of our approach in improving trajectory control. We discuss some ablations in the main text, and defer the rest to the supplementary material. Following prior work~\cite{peekaboo, trailblazer}, our experiments are conducted on Zeroscope, a widely used open-source text-to-video diffusion model. Zeroscope generates 24-frame videos at a resolution of 576$\times$320, conditioned on text prompts. We compare our proposed components—Mask Normalization and (Temporal) Intrinsic Denoising—against baseline methods including Peekaboo~\cite{peekaboo} and Trailblazer~\cite{trailblazer}. Unless otherwise specified, we set $\gamma = 0.05$ and $M = 2$ for all experiments. We evaluate our results on the randomly generated bounding boxes, as proposed by \citet{peekaboo}. The details of the dataset along with the list of prompts is available in the supplementary material. All the experiments were performed on 40GB NVIDIA A100 GPU.



\begin{table}[h]
\centering
\small
\begin{tabular}{lcccccccc}
\toprule
\multirow{2}{*}{{Model}} & \multirow{2}{*}{} & \multicolumn{3}{c}{{Static}} &\multirow{2}{*}{} & \multicolumn{3}{c}{{Dynamic}} \\
\cmidrule{3-5}
\cmidrule{7-9}
 && CLIP-SIM$\uparrow$ & CoV$\uparrow$ & mIoU$\uparrow$ && CLIP-SIM$\uparrow$ & CoV$\uparrow$ & mIoU$\uparrow$ \\
\midrule
Peekaboo \cite{peekaboo} && $31.47$ &	\colorbox{pastelblue}{$40$} &	\colorbox{pastelblue}{$29.92\%$} && $31.55$ &	$45$	& $29.76\%$ \\
Trailblazer \cite{trailblazer} && $31.50$ &	$35$& $21.18\%$ && $30.03$ &	$42$	& \colorbox{pastelgreen}{$36.01\%$} \\
\midrule
Mask Norm  && \colorbox{pastelblue}{$31.66$}	& $38$	&  $22.09\%$ &&  $30.64$ &	$32$	& $16.27\%$  \\
Mask Norm + ID && \colorbox{pastelgreen}{$32.06$} &	$38$ & $25.97\%$  &&  \colorbox{pastelgreen}{$32.17$} &	\colorbox{pastelgreen}{$49$}	& $22.32\%$  \\
Mask Norm + TID  &&  $31.48$ &	\colorbox{pastelgreen}{$41$}	& \colorbox{pastelgreen}{$33.82\%$}	 &&  \colorbox{pastelblue}{$31.80$} &	\colorbox{pastelblue}{$46$}	& \colorbox{pastelblue}{$34.73\%$}  \\
\bottomrule
\\
\end{tabular}
\caption{
\small \textbf{Conditional evaluation}: Evaluation results on static and dynamic subsets, measured in terms of CLIP-SIM~\cite{radford2021learning}, CoV, and percentage mIoU. The best result in each column is highlighted in {green}, and the second-best in {blue}. While existing methods exhibit trade-offs between CLIP-SIM, CoV, and mIoU, our proposed Mask Norm+TID method achieves strong performance consistently.}
\label{tab:cond-eval}
\end{table}

\begin{table}[h]
\centering
\small
\begin{tabular}{lccc}
\toprule
{Model} & {{FID$\downarrow$}} & {{KID$\downarrow$}} & {\color{gray} {JeDi$^{\dagger}\downarrow$}} \\
& & & \\
\midrule
Peekaboo~\cite{peekaboo} & $134.30$ & $2.03\% \pm 0.07$ &  \colorbox{pastelblue}{\color{gray} $1.40$}  \\
Trailblazer~\cite{trailblazer} & $140.07$	& $2.18\% \pm 0.07$ & {\color{gray} $1.65$}  \\
\midrule
Mask Norm  & $151.99$ &	$2.77\%\pm 0.07$  & {\color{gray}  $1.46$} \\
Mask Norm + ID  & \colorbox{pastelblue}{$133.74$} &	\colorbox{pastelgreen}{$1.81\% \pm 0.06$} &  \colorbox{pastelgreen}{\color{gray} $1.34$}  \\
Mask Norm + TID  & \colorbox{pastelgreen}{$131.19$} & \colorbox{pastelblue}{$1.92\% \pm 0.06$} & {\color{gray} $1.86$} \\
\bottomrule 
\\
\end{tabular}
\caption{ \small \textbf{Unconditional evaluation}: Evaluation results measured in terms of FID~\cite{heusel2017gans}, KID~\cite{bińkowski2018demystifying} and JeDi~\cite{luo2025beyond}.  The best result in each column is highlighted in {green}, and the second-best in {blue}. ID and TID refer to intrinsic denoising and temporal intrinsic denoising respectively. We find that Mask Norm with TID achieves consistently high FID and KID scores. $^{\dagger}$The JeDi score for other approaches is deceptively low and does not fully represent our performance, as highlighted in the main text.}
\label{tab:uncond-eval}
\end{table}

\subsection{Results and Discussion}

    We divide the evaluation metrics into two categories: conditional and unconditional. For conditional evaluation, we use the CLIP Similarity (CLIP-SIM) score~\cite{radford2021learning} to assess semantic relevance to the input prompt. To evaluate trajectory control, we adopt the strategy used by Peekaboo's paper \cite{peekaboo}. Specifically, we extract bounding boxes from generated videos using OWL-ViT~\cite{minderer2022simple} and compute the fraction of videos in which the object is detected in at least 12 frames (Coverage, or CoV). To quantify the accuracy of the predicted trajectory, we compute the mean Intersection over Union (mIoU) with respect to the input bounding boxes. For unconditional quality evaluation, we report Fr\'echet Inception Distance (FID)~\cite{heusel2017gans} and  Kernel Inception Distance (KID)~\cite{bińkowski2018demystifying} for for per-frame quality evaluation, and JeDi Score~\cite{luo2025beyond} for video quality evaluation. We use JeDi Score instead of Fr\'echet Video Distance (FVD) because of FVD is reliable only for large number of samples~\cite{luo2025beyond}. Following \citet{trailblazer}, we use the AnimalKingdom~\cite{ng2022animal} dataset as the reference set for all unconditional metrics.


    The results for conditional and unconditional evaluation are reported in \cref{tab:cond-eval} and \cref{tab:uncond-eval}, respectively. In \cref{tab:cond-eval}, we observe that Peekaboo \cite{peekaboo} achieves high coverage but suffers from low mIoU. It indicates that while the subject is often present, \cite{peekaboo} model struggles to aptly control its trajectory. In contrast, Trailblazer \cite{trailblazer} 
    performs low on static bounding boxes yet exhibits strong trajectory control for dynamic ones, albeit at the cost of semantic fidelity, as reflected in its lower CLIP-SIM scores. 

    Our method, i.e., mask normalization (Mask Norm) with intrinsic denoising (ID) and temporal  intrinsic denoising (TID), consistently performs well on all metrics: high CLIP-SIM, Coverage, and mIoU, indicating effective trajectory control without loosing semantic quality. This is further supported by results in \cref{tab:uncond-eval}, where our method consistently achieves low FID and KID, reflecting strong perceptual quality.

    Note that the low JeDi scores reported by Peekaboo \cite{peekaboo} and mask normalization with ID are misleading, as they come at the cost of significantly lower mIoU. In other words, these models are not conditioning on the trajectory, which is the core objective of this task, and instead rely mostly on text-based video generation. A similar issue has been discussed by \citet{trailblazer} and \citet{freeinit}, who highlight that because the evaluation set includes unnatural motions to increase task difficulty, the generations often deviate from the distribution of natural videos. This in turn inflates video-based perceptual scores, underscoring the importance of image-based perceptual evaluation in this setting.

\begin{figure}
\centering
\includegraphics[width=\textwidth]{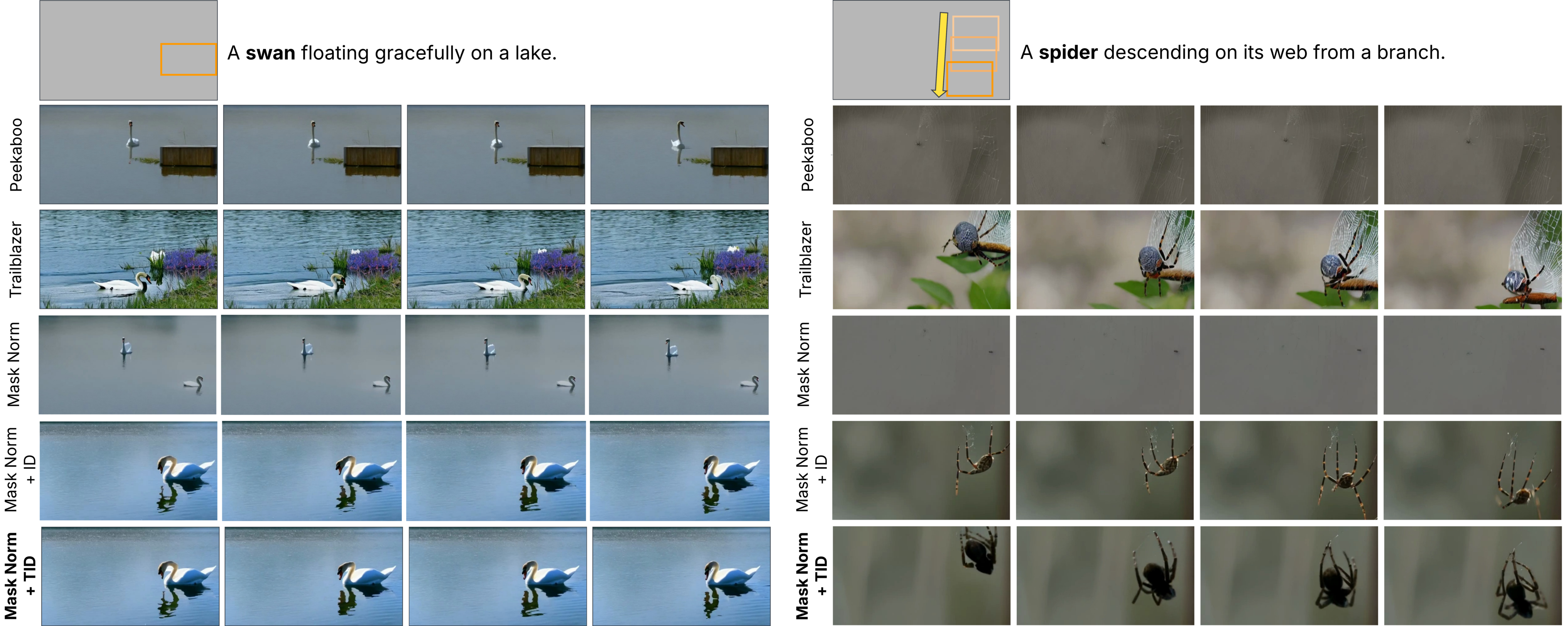}
\caption{\small \label{fig:swan-spider} \textbf{Qualitative comparison and ablations.} Peekaboo (top row) exhibits prominent artifacts in both static (left panel) and dynamic bounding box (right panel) settings. Trailblazer (2nd row) struggles with realism and often ignores the spatial constraint. Mask normalization alone (3rd row) fails to capture both perceptual quality and trajectory alignment. Adding intrinsic denoising (4th row) improves visual fidelity but still lacks precise control. Our full method, mask normalization with temporal intrinsic denoising (bottom row), produces realistic, trajectory-aligned videos. }
\end{figure}

Qualitatively, as shown in \cref{fig:swan-spider}, our method, mask normalization with temporal intrinsic denoising, produces realistic, trajectory-aligned videos with strong spatio-temporal coherence. This is not the case for Peekaboo \cite{peekaboo} and Trailblazer \cite{trailblazer}. As discussed earlier, Trailblazer for static bounding boxes, while Peekaboo struggles with both perceptual quality and control. Although Trailblazer achieves decent control for dynamic bounding boxes at high guidance scale of $11$ (as recommended by the authors), the outputs tend to be overly saturated and unnatural, which is indicative of mode collapse and is also reflected in its poor FID and KID scores.  The success of our method depends on all three components: mask normalization, temporal prior, and intrinsic denoising. As shown in \cref{fig:swan-spider}, using mask normalization alone leads to both perceptual degradation and poor trajectory control. Adding intrinsic denoising significantly improves spatial alignment, especially in cases with static bounding boxes. Finally, combining all three components results in stronger spatio-temporal control and improved perceptual quality. We provide a more detailed comparative and ablation study in the supplementary material. 

\section{Conclusion}
\label{sec:conclusion}


To conclude, this work offers a novel and principled approach to trajectory-controlled video diffusion through the lens of domain adaptation. By identifying covariate shift as a core failure mode of masked attention mechanisms, we introduce mask normalization and prior-guided intrinsic denoising that encapsulates lightweight, modality-agnostic strategies that address temporal inconsistency while preserving spatial fidelity. Rooted in domain adaptation, our methods deliver consistent improvements in interactive video generation, with minimal architectural overhead and without any retraining. Similar to previous works, our study is limited to single-subject, static-camera scenarios, the conceptual tools we introduce lay the groundwork for broader extensions to multi-agent systems and simultaneous camera control. We also underscore the limitations of current evaluation protocols and advocate for more holistic benchmarks and metrics that reflect the nuanced challenges of IVG. 

Since our method can be applied to any video diffusion model without major modifications, it inherits the broader societal limitations associated with such models. We emphasize that these systems should be used responsibly and with care so as not to violate copyright or piracy laws, nor be employed to generate harmful or malicious content.


\bibliographystyle{abbrvnat}
\bibliography{main.bib}

\begin{thebibliography}{66}
\providecommand{\natexlab}[1]{#1}
\providecommand{\url}[1]{\texttt{#1}}
\expandafter\ifx\csname urlstyle\endcsname\relax
  \providecommand{\doi}[1]{doi: #1}\else
  \providecommand{\doi}{doi: \begingroup \urlstyle{rm}\Url}\fi

\bibitem[Agarwal et~al.(2023)Agarwal, Karanam, Joseph, Saxena, Goswami, and Srinivasan]{agarwal2023star}
A.~Agarwal, S.~Karanam, K.~Joseph, A.~Saxena, K.~Goswami, and B.~V. Srinivasan.
\newblock A-star: Test-time attention segregation and retention for text-to-image synthesis.
\newblock In \emph{Proceedings of the IEEE/CVF International Conference on Computer Vision}, pages 2283--2293, 2023.

\bibitem[Bansal et~al.(2024)Bansal, Chu, Schwarzschild, Sengupta, Goldblum, Geiping, and Goldstein]{bansal2024universal}
A.~Bansal, H.-M. Chu, A.~Schwarzschild, R.~Sengupta, M.~Goldblum, J.~Geiping, and T.~Goldstein.
\newblock Universal guidance for diffusion models.
\newblock In \emph{ICLR}, 2024.
\newblock URL \url{https://openreview.net/forum?id=pzpWBbnwiJ}.

\bibitem[Bińkowski et~al.(2018)Bińkowski, Sutherland, Arbel, and Gretton]{bińkowski2018demystifying}
M.~Bińkowski, D.~J. Sutherland, M.~Arbel, and A.~Gretton.
\newblock Demystifying {MMD} {GAN}s.
\newblock In \emph{International Conference on Learning Representations}, 2018.
\newblock URL \url{https://openreview.net/forum?id=r1lUOzWCW}.

\bibitem[Brooks et~al.(2024)Brooks, Peebles, Holmes, DePue, Guo, Jing, Schnurr, Taylor, Luhman, Luhman, et~al.]{brooks2024video}
T.~Brooks, B.~Peebles, C.~Holmes, W.~DePue, Y.~Guo, L.~Jing, D.~Schnurr, J.~Taylor, T.~Luhman, E.~Luhman, et~al.
\newblock Video generation models as world simulators.
\newblock \emph{OpenAI Blog}, 1:\penalty0 8, 2024.

\bibitem[Cao et~al.(2023)Cao, Wang, Qi, Shan, Qie, and Zheng]{cao2023masactrl}
M.~Cao, X.~Wang, Z.~Qi, Y.~Shan, X.~Qie, and Y.~Zheng.
\newblock Masactrl: Tuning-free mutual self-attention control for consistent image synthesis and editing.
\newblock In \emph{Proceedings of the IEEE/CVF international conference on computer vision}, pages 22560--22570, 2023.

\bibitem[Chen et~al.(2024)Chen, Shu, Chen, He, Wang, and Li]{motion-zero}
C.~Chen, J.~Shu, L.~Chen, G.~He, C.~Wang, and Y.~Li.
\newblock Motion-zero: Zero-shot moving object control framework for diffusion-based video generation.
\newblock \emph{arXiv preprint arXiv:2401.10150}, 2024.

\bibitem[Chen et~al.(2023{\natexlab{a}})Chen, Lin, Tseng, Lin, and Yang]{chen2023motion}
T.-S. Chen, C.~H. Lin, H.-Y. Tseng, T.-Y. Lin, and M.-H. Yang.
\newblock Motion-conditioned diffusion model for controllable video synthesis.
\newblock \emph{arXiv preprint arXiv:2304.14404}, 2023{\natexlab{a}}.

\bibitem[Chen et~al.(2023{\natexlab{b}})Chen, Ji, Wu, Wu, Xie, Li, Xia, Xiao, and Lin]{chen2023control}
W.~Chen, Y.~Ji, J.~Wu, H.~Wu, P.~Xie, J.~Li, X.~Xia, X.~Xiao, and L.~Lin.
\newblock Control-a-video: Controllable text-to-video generation with diffusion models.
\newblock \emph{arXiv preprint arXiv:2305.13840}, 2023{\natexlab{b}}.

\bibitem[Coltuc et~al.(2006)Coltuc, Bolon, and Chassery]{coltuc2006exact}
D.~Coltuc, P.~Bolon, and J.-M. Chassery.
\newblock Exact histogram specification.
\newblock \emph{IEEE TIP}, 15\penalty0 (5):\penalty0 1143--1152, 2006.

\bibitem[Dhariwal and Nichol(2021)]{dhariwal2021diffusion}
P.~Dhariwal and A.~Q. Nichol.
\newblock Diffusion models beat {GAN}s on image synthesis.
\newblock In A.~Beygelzimer, Y.~Dauphin, P.~Liang, and J.~W. Vaughan, editors, \emph{NeurIPS}, 2021.
\newblock URL \url{https://openreview.net/forum?id=AAWuCvzaVt}.

\bibitem[Efron(2011)]{efron2011tweedie}
B.~Efron.
\newblock Tweedie’s formula and selection bias.
\newblock \emph{J. Am. Stat. Assoc.}, 106\penalty0 (496):\penalty0 1602--1614, 2011.

\bibitem[Epstein et~al.(2023)Epstein, Jabri, Poole, Efros, and Holynski]{epstein2023diffusion}
D.~Epstein, A.~Jabri, B.~Poole, A.~Efros, and A.~Holynski.
\newblock Diffusion self-guidance for controllable image generation.
\newblock \emph{Advances in Neural Information Processing Systems}, 36:\penalty0 16222--16239, 2023.

\bibitem[Esser et~al.(2023)Esser, Chiu, Atighehchian, Granskog, and Germanidis]{esser2023structure}
P.~Esser, J.~Chiu, P.~Atighehchian, J.~Granskog, and A.~Germanidis.
\newblock Structure and content-guided video synthesis with diffusion models.
\newblock In \emph{Proceedings of the IEEE/CVF international conference on computer vision}, pages 7346--7356, 2023.

\bibitem[Geng and Owens(2024)]{gengmotion}
D.~Geng and A.~Owens.
\newblock Motion guidance: Diffusion-based image editing with differentiable motion estimators.
\newblock In \emph{The Twelfth International Conference on Learning Representations}, 2024.

\bibitem[Goodfellow et~al.(2014)Goodfellow, Pouget-Abadie, Mirza, Xu, Warde-Farley, Ozair, Courville, and Bengio]{goodfellow2014generative}
I.~Goodfellow, J.~Pouget-Abadie, M.~Mirza, B.~Xu, D.~Warde-Farley, S.~Ozair, A.~Courville, and Y.~Bengio.
\newblock Generative adversarial nets.
\newblock \emph{Advances in neural information processing systems}, 27, 2014.

\bibitem[Heusel et~al.(2017)Heusel, Ramsauer, Unterthiner, Nessler, and Hochreiter]{heusel2017gans}
M.~Heusel, H.~Ramsauer, T.~Unterthiner, B.~Nessler, and S.~Hochreiter.
\newblock Gans trained by a two time-scale update rule converge to a local nash equilibrium.
\newblock \emph{Advances in neural information processing systems}, 30, 2017.

\bibitem[Ho et~al.(2022)Ho, Salimans, Gritsenko, Chan, Norouzi, and Fleet]{ho2022video}
J.~Ho, T.~Salimans, A.~Gritsenko, W.~Chan, M.~Norouzi, and D.~J. Fleet.
\newblock Video diffusion models.
\newblock \emph{Advances in Neural Information Processing Systems}, 35:\penalty0 8633--8646, 2022.

\bibitem[Hu et~al.(2021)Hu, Shen, Wallis, Allen-Zhu, Li, Wang, Wang, and Chen]{hu2021lora}
E.~J. Hu, Y.~Shen, P.~Wallis, Z.~Allen-Zhu, Y.~Li, S.~Wang, L.~Wang, and W.~Chen.
\newblock Lora: Low-rank adaptation of large language models.
\newblock \emph{arXiv preprint arXiv:2106.09685}, 2021.

\bibitem[Hu et~al.(2023)Hu, Chen, and Luo]{hu2023lamd}
Y.~Hu, Z.~Chen, and C.~Luo.
\newblock Lamd: Latent motion diffusion for video generation.
\newblock \emph{arXiv preprint arXiv:2304.11603}, 2023.

\bibitem[Huang et~al.(2023)Huang, Feng, Shi, Xu, Yu, and Yang]{huang2023free}
H.~Huang, Y.~Feng, C.~Shi, L.~Xu, J.~Yu, and S.~Yang.
\newblock Free-bloom: Zero-shot text-to-video generator with llm director and ldm animator.
\newblock \emph{Advances in Neural Information Processing Systems}, 36:\penalty0 26135--26158, 2023.

\bibitem[Jain et~al.(2024)Jain, Nasery, Vineet, and Behl]{peekaboo}
Y.~Jain, A.~Nasery, V.~Vineet, and H.~Behl.
\newblock Peekaboo: Interactive video generation via masked-diffusion.
\newblock In \emph{CVPR}, pages 8079--8088, 2024.

\bibitem[Jeong et~al.(2024{\natexlab{a}})Jeong, Chang, Park, and Ye]{jeong2024dreammotion}
H.~Jeong, J.~Chang, G.~Y. Park, and J.~C. Ye.
\newblock Dreammotion: Space-time self-similar score distillation for zero-shot video editing.
\newblock In \emph{European Conference on Computer Vision}, pages 358--376. Springer, 2024{\natexlab{a}}.

\bibitem[Jeong et~al.(2024{\natexlab{b}})Jeong, Park, and Ye]{jeong2024vmc}
H.~Jeong, G.~Y. Park, and J.~C. Ye.
\newblock Vmc: Video motion customization using temporal attention adaption for text-to-video diffusion models.
\newblock In \emph{Proceedings of the IEEE/CVF Conference on Computer Vision and Pattern Recognition}, pages 9212--9221, 2024{\natexlab{b}}.

\bibitem[Khachatryan et~al.(2023)Khachatryan, Movsisyan, Tadevosyan, Henschel, Wang, Navasardyan, and Shi]{khachatryan2023text2video}
L.~Khachatryan, A.~Movsisyan, V.~Tadevosyan, R.~Henschel, Z.~Wang, S.~Navasardyan, and H.~Shi.
\newblock Text2video-zero: Text-to-image diffusion models are zero-shot video generators.
\newblock In \emph{Proceedings of the IEEE/CVF International Conference on Computer Vision}, pages 15954--15964, 2023.

\bibitem[Li et~al.(2024)Li, Wan, Moens, and Tuytelaars]{li2024animate}
M.~Li, B.~Wan, M.-F. Moens, and T.~Tuytelaars.
\newblock Animate your motion: Turning still images into dynamic videos.
\newblock In \emph{European Conference on Computer Vision}, pages 409--425. Springer, 2024.

\bibitem[Li et~al.(2025)Li, Lai, Xu, Qu, Cao, Zhang, Dai, and Ji]{li2025director3d}
X.~Li, Z.~Lai, L.~Xu, Y.~Qu, L.~Cao, S.~Zhang, B.~Dai, and R.~Ji.
\newblock Director3d: Real-world camera trajectory and 3d scene generation from text.
\newblock \emph{Advances in Neural Information Processing Systems}, 37:\penalty0 75125--75151, 2025.

\bibitem[Li et~al.(2023)Li, Liu, Wu, Mu, Yang, Gao, Li, and Lee]{li2023gligen}
Y.~Li, H.~Liu, Q.~Wu, F.~Mu, J.~Yang, J.~Gao, C.~Li, and Y.~J. Lee.
\newblock Gligen: Open-set grounded text-to-image generation.
\newblock In \emph{Proceedings of the IEEE/CVF conference on computer vision and pattern recognition}, pages 22511--22521, 2023.

\bibitem[Lin et~al.(2024)Lin, Liu, Li, and Yang]{lin2024common}
S.~Lin, B.~Liu, J.~Li, and X.~Yang.
\newblock Common diffusion noise schedules and sample steps are flawed.
\newblock In \emph{Proceedings of the IEEE/CVF winter conference on applications of computer vision}, pages 5404--5411, 2024.

\bibitem[Liu et~al.(2024)Liu, Zhang, Li, Lin, and Jia]{liu2024video}
S.~Liu, Y.~Zhang, W.~Li, Z.~Lin, and J.~Jia.
\newblock Video-p2p: Video editing with cross-attention control.
\newblock In \emph{Proceedings of the IEEE/CVF Conference on Computer Vision and Pattern Recognition}, pages 8599--8608, 2024.

\bibitem[Lugmayr et~al.(2022)Lugmayr, Danelljan, Romero, Yu, Timofte, and Van~Gool]{lugmayr2022repaint}
A.~Lugmayr, M.~Danelljan, A.~Romero, F.~Yu, R.~Timofte, and L.~Van~Gool.
\newblock Repaint: Inpainting using denoising diffusion probabilistic models.
\newblock In \emph{Proceedings of the IEEE/CVF conference on computer vision and pattern recognition}, pages 11461--11471, 2022.

\bibitem[Luo et~al.(2025)Luo, Favero, Luo, Jolicoeur-Martineau, and Pal]{luo2025beyond}
G.~Y. Luo, G.~M. Favero, Z.~Luo, A.~Jolicoeur-Martineau, and C.~Pal.
\newblock Beyond {FVD}: An enhanced evaluation metrics for video generation distribution quality.
\newblock In \emph{The Thirteenth International Conference on Learning Representations}, 2025.
\newblock URL \url{https://openreview.net/forum?id=cC3LxGZasH}.

\bibitem[Ma et~al.(2024)Ma, Lewis, and Kleijn]{trailblazer}
W.-D.~K. Ma, J.~P. Lewis, and W.~B. Kleijn.
\newblock Trailblazer: Trajectory control for diffusion-based video generation.
\newblock In \emph{SIGGRAPH Asia}, pages 1--11, 2024.

\bibitem[Minderer et~al.(2022)Minderer, Gritsenko, Stone, Neumann, Weissenborn, Dosovitskiy, Mahendran, Arnab, Dehghani, Shen, et~al.]{minderer2022simple}
M.~Minderer, A.~Gritsenko, A.~Stone, M.~Neumann, D.~Weissenborn, A.~Dosovitskiy, A.~Mahendran, A.~Arnab, M.~Dehghani, Z.~Shen, et~al.
\newblock Simple open-vocabulary object detection.
\newblock In \emph{European conference on computer vision}, pages 728--755. Springer, 2022.

\bibitem[Monge(1781)]{monge1781memoire}
G.~Monge.
\newblock M{\'e}moire sur la th{\'e}orie des d{\'e}blais et des remblais.
\newblock \emph{Mem. Math. Phys. Acad. Royale Sci.}, pages 666--704, 1781.

\bibitem[Ng et~al.(2022)Ng, Ong, Zheng, Ni, Yeo, and Liu]{ng2022animal}
X.~L. Ng, K.~E. Ong, Q.~Zheng, Y.~Ni, S.~Y. Yeo, and J.~Liu.
\newblock Animal kingdom: A large and diverse dataset for animal behavior understanding.
\newblock In \emph{Proceedings of the IEEE/CVF conference on computer vision and pattern recognition}, pages 19023--19034, 2022.

\bibitem[{Nicki Skafte Detlefsen} et~al.(2022){Nicki Skafte Detlefsen}, {Jiri Borovec}, {Justus Schock}, {Ananya Harsh}, {Teddy Koker}, {Luca Di Liello}, {Daniel Stancl}, {Changsheng Quan}, {Maxim Grechkin}, and {William Falcon}]{torchmertrics}
{Nicki Skafte Detlefsen}, {Jiri Borovec}, {Justus Schock}, {Ananya Harsh}, {Teddy Koker}, {Luca Di Liello}, {Daniel Stancl}, {Changsheng Quan}, {Maxim Grechkin}, and {William Falcon}.
\newblock {TorchMetrics - Measuring Reproducibility in PyTorch}, Feb. 2022.
\newblock URL \url{https://github.com/Lightning-AI/torchmetrics}.

\bibitem[Niu et~al.(2024)Niu, Cun, Wang, Zhang, Shan, and Zheng]{niu2024mofa}
M.~Niu, X.~Cun, X.~Wang, Y.~Zhang, Y.~Shan, and Y.~Zheng.
\newblock Mofa-video: Controllable image animation via generative motion field adaptions in frozen image-to-video diffusion model.
\newblock In \emph{European Conference on Computer Vision}, pages 111--128. Springer, 2024.

\bibitem[Phung et~al.(2024)Phung, Ge, and Huang]{phung2024grounded}
Q.~Phung, S.~Ge, and J.-B. Huang.
\newblock Grounded text-to-image synthesis with attention refocusing.
\newblock In \emph{Proceedings of the IEEE/CVF Conference on Computer Vision and Pattern Recognition}, pages 7932--7942, 2024.

\bibitem[Qiu et~al.(2024{\natexlab{a}})Qiu, Chen, Wang, He, Xia, and Liu]{freetraj}
H.~Qiu, Z.~Chen, Z.~Wang, Y.~He, M.~Xia, and Z.~Liu.
\newblock Freetraj: Tuning-free trajectory control in video diffusion models.
\newblock \emph{arXiv preprint arXiv:2406.16863}, 2024{\natexlab{a}}.

\bibitem[Qiu et~al.(2024{\natexlab{b}})Qiu, Xia, Zhang, He, Wang, Shan, and Liu]{qiufreenoise}
H.~Qiu, M.~Xia, Y.~Zhang, Y.~He, X.~Wang, Y.~Shan, and Z.~Liu.
\newblock Freenoise: Tuning-free longer video diffusion via noise rescheduling.
\newblock In \emph{The Twelfth International Conference on Learning Representations}, 2024{\natexlab{b}}.

\bibitem[Radford et~al.(2021)Radford, Kim, Hallacy, Ramesh, Goh, Agarwal, Sastry, Askell, Mishkin, Clark, et~al.]{radford2021learning}
A.~Radford, J.~W. Kim, C.~Hallacy, A.~Ramesh, G.~Goh, S.~Agarwal, G.~Sastry, A.~Askell, P.~Mishkin, J.~Clark, et~al.
\newblock Learning transferable visual models from natural language supervision.
\newblock In \emph{International conference on machine learning}, pages 8748--8763. PMLR, 2021.

\bibitem[Ren et~al.(2024)Ren, Zhou, Yang, Shi, Liu, Liu, Kwon, and Shrivastava]{ren2024customize}
Y.~Ren, Y.~Zhou, J.~Yang, J.~Shi, D.~Liu, F.~Liu, M.~Kwon, and A.~Shrivastava.
\newblock Customize-a-video: One-shot motion customization of text-to-video diffusion models.
\newblock In \emph{European Conference on Computer Vision}, pages 332--349. Springer, 2024.

\bibitem[Rolland et~al.(2000)Rolland, Vo, Bloss, and Abbey]{rolland2000fast}
J.~P. Rolland, V.~Vo, B.~Bloss, and C.~K. Abbey.
\newblock Fast algorithms for histogram matching: Application to texture synthesis.
\newblock \emph{Journal of Electronic Imaging}, 9\penalty0 (1):\penalty0 39--45, 2000.

\bibitem[Ronneberger et~al.(2015)Ronneberger, Fischer, and Brox]{ronneberger2015u}
O.~Ronneberger, P.~Fischer, and T.~Brox.
\newblock U-net: Convolutional networks for biomedical image segmentation.
\newblock In \emph{Medical image computing and computer-assisted intervention--MICCAI 2015: 18th international conference, Munich, Germany, October 5-9, 2015, proceedings, part III 18}, pages 234--241. Springer, 2015.

\bibitem[Singer et~al.(2022)Singer, Polyak, Hayes, Yin, An, Zhang, Hu, Yang, Ashual, Gafni, et~al.]{singer2022make}
U.~Singer, A.~Polyak, T.~Hayes, X.~Yin, J.~An, S.~Zhang, Q.~Hu, H.~Yang, O.~Ashual, O.~Gafni, et~al.
\newblock Make-a-video: Text-to-video generation without text-video data.
\newblock \emph{arXiv preprint arXiv:2209.14792}, 2022.

\bibitem[Ulyanov et~al.(2018)Ulyanov, Vedaldi, and Lempitsky]{deepimageprior}
D.~Ulyanov, A.~Vedaldi, and V.~Lempitsky.
\newblock Deep image prior.
\newblock In \emph{CVPR}, pages 9446--9454, 2018.

\bibitem[Wang et~al.()Wang, Zhang, Zou, Zeng, Wei, Yuan, and Li]{wangboximator}
J.~Wang, Y.~Zhang, J.~Zou, Y.~Zeng, G.~Wei, L.~Yuan, and H.~Li.
\newblock Boximator: Generating rich and controllable motions for video synthesis.
\newblock In \emph{Forty-first International Conference on Machine Learning (ICML)}.

\bibitem[Wang et~al.(2023)Wang, Yuan, Zhang, Chen, Wang, Zhang, Shen, Zhao, and Zhou]{wang2023videocomposer}
X.~Wang, H.~Yuan, S.~Zhang, D.~Chen, J.~Wang, Y.~Zhang, Y.~Shen, D.~Zhao, and J.~Zhou.
\newblock Videocomposer: Compositional video synthesis with motion controllability.
\newblock \emph{Advances in Neural Information Processing Systems}, 36:\penalty0 7594--7611, 2023.

\bibitem[Wang et~al.(2024)Wang, Yuan, Wang, Li, Chen, Xia, Luo, and Shan]{wang2024motionctrl}
Z.~Wang, Z.~Yuan, X.~Wang, Y.~Li, T.~Chen, M.~Xia, P.~Luo, and Y.~Shan.
\newblock Motionctrl: A unified and flexible motion controller for video generation.
\newblock In \emph{ACM SIGGRAPH 2024 Conference Papers}, pages 1--11, 2024.

\bibitem[Wei et~al.(2024)Wei, Zhang, Qing, Yuan, Liu, Liu, Zhang, Zhou, and Shan]{wei2024dreamvideo}
Y.~Wei, S.~Zhang, Z.~Qing, H.~Yuan, Z.~Liu, Y.~Liu, Y.~Zhang, J.~Zhou, and H.~Shan.
\newblock Dreamvideo: Composing your dream videos with customized subject and motion.
\newblock In \emph{Proceedings of the IEEE/CVF Conference on Computer Vision and Pattern Recognition}, pages 6537--6549, 2024.

\bibitem[Wu et~al.(2023)Wu, Ge, Wang, Lei, Gu, Shi, Hsu, Shan, Qie, and Shou]{wu2023tune}
J.~Z. Wu, Y.~Ge, X.~Wang, S.~W. Lei, Y.~Gu, Y.~Shi, W.~Hsu, Y.~Shan, X.~Qie, and M.~Z. Shou.
\newblock Tune-a-video: One-shot tuning of image diffusion models for text-to-video generation.
\newblock In \emph{Proceedings of the IEEE/CVF International Conference on Computer Vision}, pages 7623--7633, 2023.

\bibitem[Wu et~al.(2024{\natexlab{a}})Wu, Chen, Yang, Guo, Li, and Zhang]{wu2024lamp}
R.~Wu, L.~Chen, T.~Yang, C.~Guo, C.~Li, and X.~Zhang.
\newblock Lamp: Learn a motion pattern for few-shot video generation.
\newblock In \emph{Proceedings of the IEEE/CVF Conference on Computer Vision and Pattern Recognition}, pages 7089--7098, 2024{\natexlab{a}}.

\bibitem[Wu et~al.(2024{\natexlab{b}})Wu, Si, Jiang, Huang, and Liu]{freeinit}
T.~Wu, C.~Si, Y.~Jiang, Z.~Huang, and Z.~Liu.
\newblock Freeinit: Bridging initialization gap in video diffusion models.
\newblock In \emph{ECCV}, pages 378--394. Springer, 2024{\natexlab{b}}.

\bibitem[Wu et~al.(2024{\natexlab{c}})Wu, Si, Jiang, Huang, and Liu]{wu2024freeinit}
T.~Wu, C.~Si, Y.~Jiang, Z.~Huang, and Z.~Liu.
\newblock Freeinit: Bridging initialization gap in video diffusion models.
\newblock In \emph{European Conference on Computer Vision}, pages 378--394. Springer, 2024{\natexlab{c}}.

\bibitem[Wu et~al.(2024{\natexlab{d}})Wu, Zhang, Wang, Zhou, Zheng, Qi, Shan, and Li]{wu2024customcrafter}
T.~Wu, Y.~Zhang, X.~Wang, X.~Zhou, G.~Zheng, Z.~Qi, Y.~Shan, and X.~Li.
\newblock Customcrafter: Customized video generation with preserving motion and concept composition abilities.
\newblock \emph{arXiv preprint arXiv:2408.13239}, 2024{\natexlab{d}}.

\bibitem[Xiao et~al.(2024{\natexlab{a}})Xiao, Feng, Zhang, Liu, Yang, Zhu, Fu, Zhu, Liu, and Zha]{dreamclean}
J.~Xiao, R.~Feng, H.~Zhang, Z.~Liu, Z.~Yang, Y.~Zhu, X.~Fu, K.~Zhu, Y.~Liu, and Z.-J. Zha.
\newblock Dreamclean: Restoring clean image using deep diffusion prior.
\newblock In \emph{ICLR}, 2024{\natexlab{a}}.
\newblock URL \url{https://openreview.net/forum?id=6ALuy19mPa}.

\bibitem[Xiao et~al.(2024{\natexlab{b}})Xiao, Zhou, Yang, and Pan]{xiao2024video}
Z.~Xiao, Y.~Zhou, S.~Yang, and X.~Pan.
\newblock Video diffusion models are training-free motion interpreter and controller.
\newblock In \emph{The Thirty-eighth Annual Conference on Neural Information Processing Systems}, 2024{\natexlab{b}}.

\bibitem[Yan et~al.(2021)Yan, Zhang, Abbeel, and Srinivas]{yan2021videogpt}
W.~Yan, Y.~Zhang, P.~Abbeel, and A.~Srinivas.
\newblock Videogpt: Video generation using vq-vae and transformers.
\newblock \emph{arXiv preprint arXiv:2104.10157}, 2021.

\bibitem[Yang et~al.(2024{\natexlab{a}})Yang, Hou, Huang, Ma, Wan, Zhang, Chen, and Liao]{direct-a-vid}
S.~Yang, L.~Hou, H.~Huang, C.~Ma, P.~Wan, D.~Zhang, X.~Chen, and J.~Liao.
\newblock Direct-a-video: Customized video generation with user-directed camera movement and object motion.
\newblock In \emph{SIGGRAPH}, pages 1--12, 2024{\natexlab{a}}.

\bibitem[Yang et~al.(2024{\natexlab{b}})Yang, Hou, Huang, Ma, Wan, Zhang, Chen, and Liao]{yang2024direct}
S.~Yang, L.~Hou, H.~Huang, C.~Ma, P.~Wan, D.~Zhang, X.~Chen, and J.~Liao.
\newblock Direct-a-video: Customized video generation with user-directed camera movement and object motion.
\newblock In \emph{ACM SIGGRAPH 2024 Conference Papers}, pages 1--12, 2024{\natexlab{b}}.

\bibitem[Zhang et~al.(2023{\natexlab{a}})Zhang, Rao, and Agrawala]{zhang2023adding}
L.~Zhang, A.~Rao, and M.~Agrawala.
\newblock Adding conditional control to text-to-image diffusion models.
\newblock In \emph{Proceedings of the IEEE/CVF International Conference on Computer Vision}, pages 3836--3847, 2023{\natexlab{a}}.

\bibitem[Zhang et~al.(2022)Zhang, Li, Li, Jia, and Zhang]{efdm}
Y.~Zhang, M.~Li, R.~Li, K.~Jia, and L.~Zhang.
\newblock Exact feature distribution matching for arbitrary style transfer and domain generalization.
\newblock In \emph{CVPR}, pages 8035--8045, 2022.

\bibitem[Zhang et~al.(2023{\natexlab{b}})Zhang, Wei, Jiang, Zhang, Zuo, and Tian]{zhang2023controlvideo}
Y.~Zhang, Y.~Wei, D.~Jiang, X.~Zhang, W.~Zuo, and Q.~Tian.
\newblock Controlvideo: Training-free controllable text-to-video generation.
\newblock \emph{arXiv preprint arXiv:2305.13077}, 2023{\natexlab{b}}.

\bibitem[Zhang et~al.(2024)Zhang, Liao, Li, Dai, Qiu, Zhu, Qin, and Wang]{zhang2024tora}
Z.~Zhang, J.~Liao, M.~Li, Z.~Dai, B.~Qiu, S.~Zhu, L.~Qin, and W.~Wang.
\newblock Tora: Trajectory-oriented diffusion transformer for video generation.
\newblock \emph{arXiv preprint arXiv:2407.21705}, 2024.

\bibitem[Zhao et~al.(2024)Zhao, Gu, Wu, Zhang, Liu, Wu, Keppo, and Shou]{zhao2024motiondirector}
R.~Zhao, Y.~Gu, J.~Z. Wu, D.~J. Zhang, J.-W. Liu, W.~Wu, J.~Keppo, and M.~Z. Shou.
\newblock Motiondirector: Motion customization of text-to-video diffusion models.
\newblock In \emph{European Conference on Computer Vision}, pages 273--290. Springer, 2024.

\bibitem[Zhou et~al.(2022)Zhou, Wang, Yan, Lv, Zhu, and Feng]{zhou2022magicvideo}
D.~Zhou, W.~Wang, H.~Yan, W.~Lv, Y.~Zhu, and J.~Feng.
\newblock Magicvideo: Efficient video generation with latent diffusion models.
\newblock \emph{arXiv preprint arXiv:2211.11018}, 2022.

\end{thebibliography}


\appendix


\newpage

\clearpage
\setcounter{page}{1}

\maketitlesupplementary


\begin{abstract}
    This supplementary material extends the empirical study of our main paper by detailing implementation settings, datasets, and additional experiments that further substantiate our claims. We first document our standardized experimental protocol: all videos are synthesized with the pre-trained Zeroscope backbone. Baselines are reproduced from the official Peekaboo \cite{peekaboo} and Trailblazer \cite{trailblazer} repositories to ensure a fair comparison. Next, we describe the evaluation dataset—the 126 prompt + bounding-box pairs introduced by Peekaboo—and outline the conditional (CLIP-SIM, Coverage, mIoU) and unconditional (FID, KID, JeDi) metrics used. We clarify our OWL-ViT filtering protocol and reference set (Animal Kingdom) for completeness. The core of this supplementary draft presents three sets of results: \textbf{\textit{(i)}} Additional qualitative results demonstrating that the synergy of mask normalization and temporal intrinsic denoising reliably enforces trajectory control while preserving visual fidelity.
    \textbf{\textit{(ii)}} Extended comparisons demonstrate that our method consistently outperforms other approaches, especially in challenging cases where baselines either lose the subject or fail to follow the prescribed motion path.
   \textbf{\textit{(iii)}} Comprehensive ablations quantifying the influence of mask normalization and the temporal-prior gradient scale. We also provide video results as part of the supplementary material.  These additional empirical studies further establish the aptness of our domain-adaptation take on the IVG problem.
\end{abstract}

\section{Experiment Details}
We use the pretrained Zeroscope model for all our experiments. All experiments are conducted on a single 40GB NVIDIA A100 GPU, generating 24 frames at a resolution of $320 \times 576$ per video with a guidance scale of 9 and 4 frozen steps. For comparisons, we use the official implementations of {Peekaboo}\footnote{\url{https://github.com/microsoft/Peekaboo}}~\cite{peekaboo} and {Trailblazer}\footnote{\url{https://github.com/hohonu-vicml/TrailBlazer}}~\cite{trailblazer}.

For all experiments involving (temporal) intrinsic denoising, we set $\gamma = 0.05$ and $M = 2$. For temporal intrinsic denoising, we use $c_g = 10000$, and for intrinsic denoising, $c_g = 0$.

\subsection{Evaluation Dataset}
We use the same set of 126 prompt–bounding box pairs as used by \citet{peekaboo}. For completeness, we provide the list of prompts below with the subject in bold:

\begin{itemize}
    \item A \textbf{red double-decker bus} moving through London streets. 
    \item A \textbf{kangaroo} standing in the Australian outback. 
    \item A \textbf{camel} resting in a desert landscape.
    \item A \textbf{deer} standing in a snowy field.
    \item An \textbf{owl} perched silently in a tree at night.
    \item A \textbf{rocket} launching into space from a launchpad.
    \item A \textbf{frog} leaping up to catch a fly.
    \item A \textbf{satellite} orbiting Earth in outer space.
    \item A red British \textbf{telephone box} on a city street.
    \item A \textbf{sailboat} gliding over the ocean waves.
    \item An old-fashioned \textbf{street lamp} on a foggy night.
    \item A \textbf{bird} diving towards the water to catch fish.
    \item A \textbf{parrot} flying upwards towards the treetops.
    \item A \textbf{squirrel} descending a tree after gathering nuts.
    \item A \textbf{duck} diving underwater in search of food.
    \item A \textbf{kangaroo} hopping down a gentle slope.
    \item A \textbf{woodpecker} climbing up a tree trunk.
    \item A \textbf{fox} sitting in a forest clearing.
    \item A \textbf{vintage car} cruising down a coastal highway.
    \item A colorful \textbf{hot air balloon} tethered to the ground.
    \item A \textbf{streetcar} trundling down tracks in a historic district.
    \item A \textbf{panda} munching bamboo in a bamboo forest.
    \item A \textbf{horse} grazing in a meadow.
    \item A \textbf{squirrel} jumping from one tree to another
    \item A \textbf{hot air balloon} drifting across a clear sky.
    \item A \textbf{leaf} falling gently from a tree.
    \item A \textbf{bear} climbing down a tree after spotting a threat.
    \item A \textbf{grand piano} in an elegant concert hall.
    \item A \textbf{vintage car} parked in front of a classic diner.
    \item A \textbf{swan} floating gracefully on a lake.
    \item A \textbf{lion} lying in the savanna grass.
    \item A \textbf{helicopter} hovering above a cityscape.
    \item A \textbf{spider} descending on its web from a branch.
    \item A classic \textbf{steam train} stationed at an old railway platform.
    \item An \textbf{owl} swooping down on its prey during the night.
    \item A \textbf{jet plane} flying high in the sky.
    \item A \textbf{penguin} standing on an iceberg.
    \item A \textbf{rabbit} burrowing downwards into its warren.
    \item A \textbf{dolphin} just breaking the ocean surface.
    \item A \textbf{skateboarder} performing tricks at a skate park.
    \item A \textbf{paper plane} gliding in the air.
    \item A \textbf{roller coaster} looping in an amusement park.
\end{itemize}
\subsection{Evaluation}
We perform conditional evaluation using CLIP-SIM, Coverage, and mIoU. Coverage is computed as the fraction of generated videos in which OWL-ViT detects the subject in at least 12 out of 24 frames (i.e., 50\%). CLIP-SIM and mIoU are then calculated on this filtered subset.

For unconditional evaluation, we use FID, KID, and JeDi scores, computed over all generated videos. The reference set is AnimalKingdom. For FID and KID, we use InceptionNet V3 as the feature extractor, using all 24 frames from each generated video and 8 uniformly sampled frames from each real video. All frames are resized to $(299, 299)$, and we use the default parameters from \texttt{torchmetrics}~\cite{torchmertrics}.

For JeDi score, we follow the official implementation\footnote{\url{https://github.com/oooolga/JEDi}}, using a batch size of 64, resized to $(224, 224)$, and averaging results over 5 random seeds.

\section{Results}

\begin{figure}
\centering
\includegraphics[width=\textwidth]{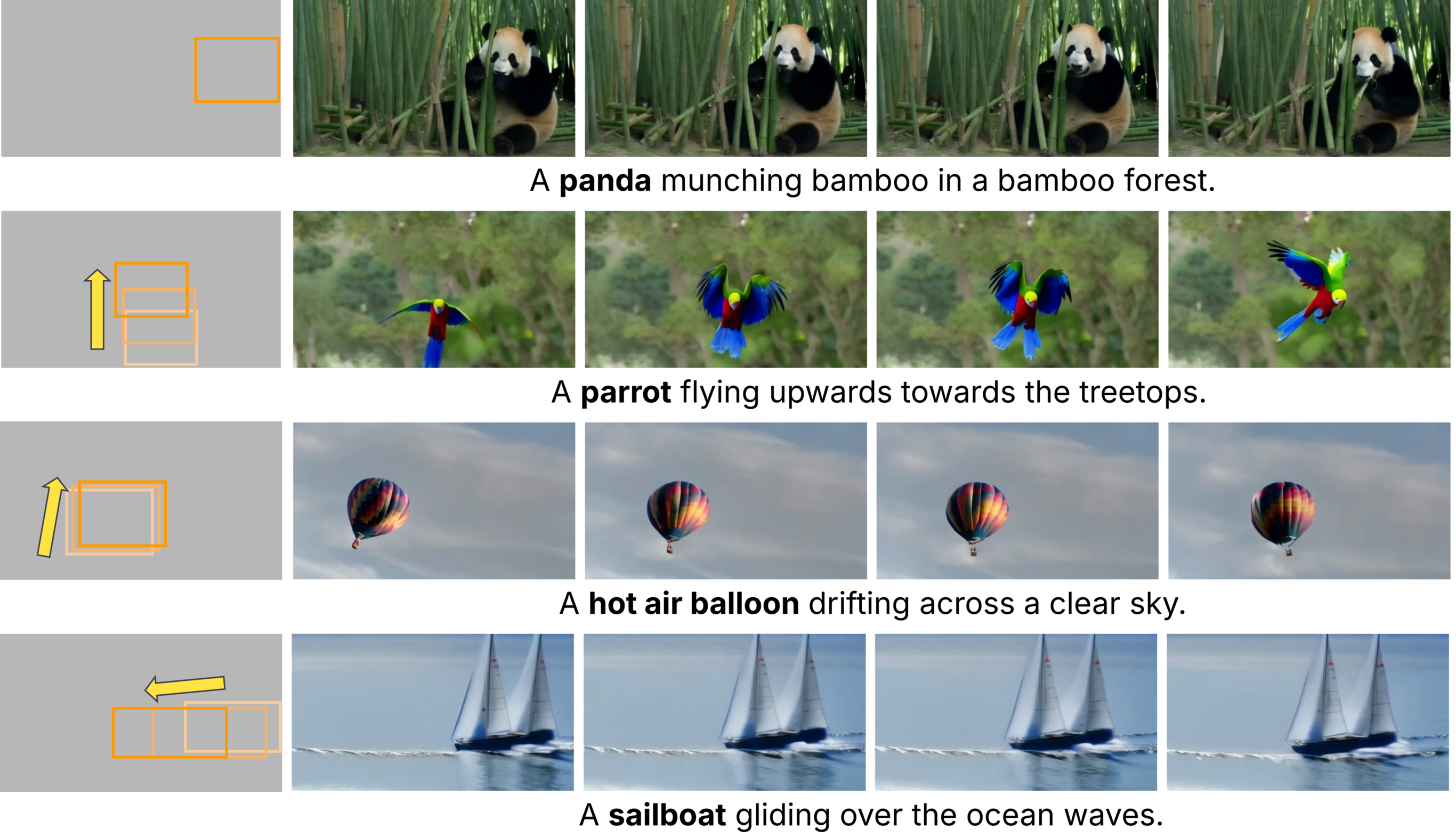}
\caption{\small \label{fig:supp-more-res} \textbf{Additional Results: } Additional examples of trajectory control in videos using mask normalization and temporal intrinsic denoising. } 
\end{figure}

We provide some more additional results demonstrating the prowess of mask normalization and temporal intrinsic denoising in trajectory control in \cref{fig:supp-more-res}

\subsection{Comparison}
\begin{figure}
\centering
\includegraphics[width=\textwidth]{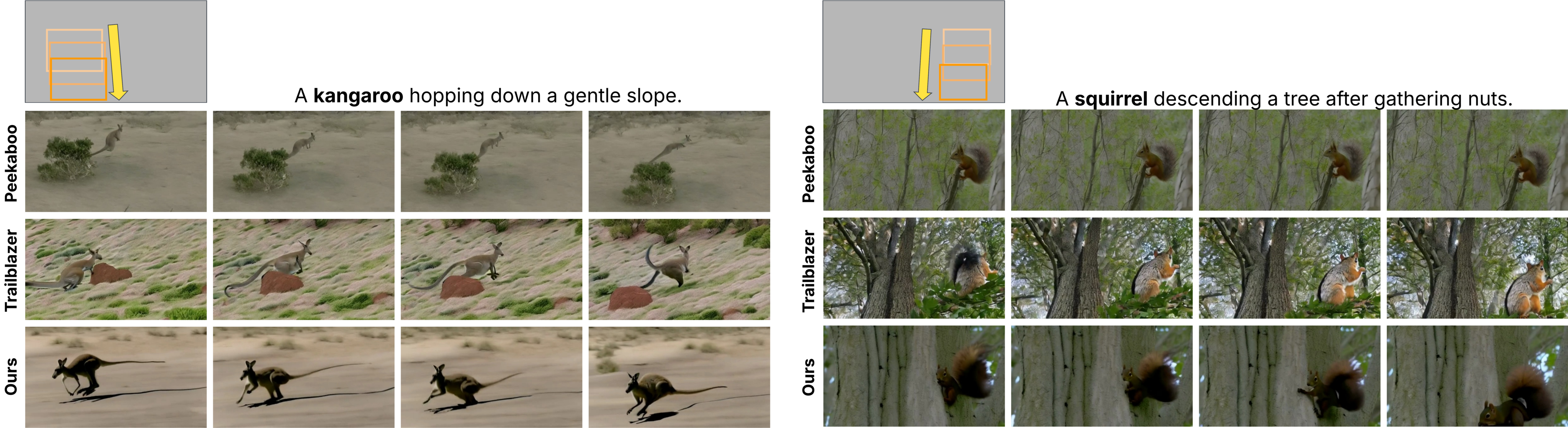}
\caption{\small \label{fig:supp-compare} \textbf{Additional Qualitative Comparison} We compare our method (mask normalization with temporal intrinsic denoising; bottom row) with Peekaboo (top row) and Trailblazer (middle row). We find that while Trailblazer and Peekaboo either fail to generate subject within the bounding boxes (kangaroo; left pane) or motion (squirrel; right pane), our method, in general, generates text-prompt and trajectory conditioned videos.} 
\end{figure}

In this subsection we present more figures comparing our results with Peekaboo and Trailblazer. As shown in \cref{fig:supp-compare}, we consistently find that mask normalization with temporal intrinsic denoising achieves much better results.

\subsection{Ablation}

\begin{figure}
\centering
\includegraphics[width=\textwidth]{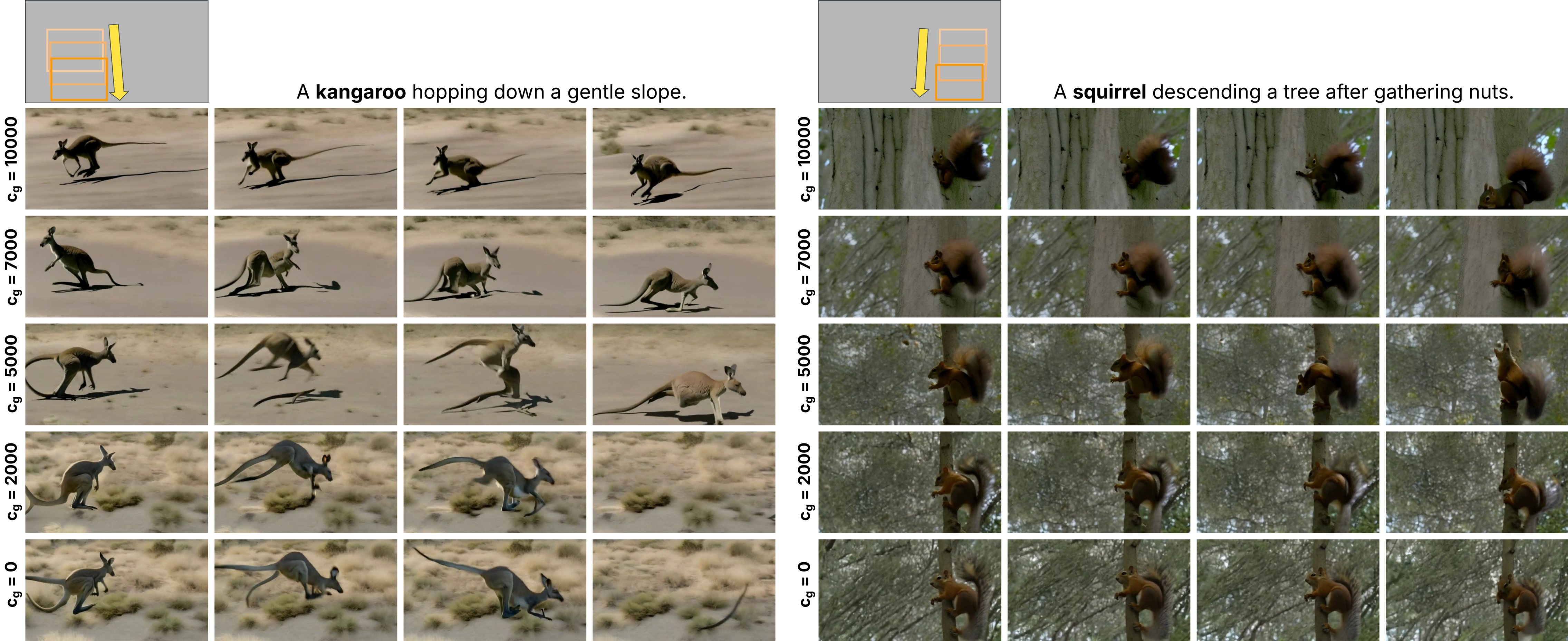}
\caption{\small \label{fig:cg-ablation} \textbf{Ablation: effect of $c_g$} We compare the effect of $c_g$ in temporal intrinsic denoising on the generated videos, for values $10000$, $7000$, $5000$, $2000$ and $0$ (corresponds to intrinsic denoising), with values decreasing along the column. Lower $c_g$ values fail to enforce spatio-temporal control.} 
\end{figure}

In this section we study the effect of mask normalization and $c_g$, the scale of the gradient from the temporal prior on the generation quality. The qualitative results are illustrated in  \cref{fig:cg-ablation}. We find that decreasing the value of $c_g$ leads to loss of trajectory control and/or introduces perceptual artifacts. We provide quantitative results for conditional and unconditional evaluation in \cref{tab:supp-cond-eval} and \cref{tab:supp-uncond-eval} that shows that mask normalization with temporal intrinsic denoising achieves the best balance between perceptual appearance and trajectory control. 

\begin{table}
\centering
\small
\begin{tabular}{cc|cccccccc}
\toprule
\multirow{2}{*}{{Mask Norm}} & \multirow{2}{*}{{$c_g$}} & \multirow{2}{*}{} & \multicolumn{3}{c}{{Static}} &\multirow{2}{*}{} & \multicolumn{3}{c}{{Dynamic}} \\
\cmidrule{4-6}
\cmidrule{8-10}
 & && CLIP-SIM$\uparrow$ & CoV$\uparrow$ & mIoU$\uparrow$ && CLIP-SIM$\uparrow$ & CoV$\uparrow$ & mIoU$\uparrow$ \\
\midrule

\checkmark  & $0$ && \colorbox{pastelgreen}{$32.06$} &	\colorbox{pastelblue}{$38$} & $25.97\%$  &&  \colorbox{pastelgreen}{$32.17$} &	\colorbox{pastelgreen}{$49$}	& $22.32\%$  \\
\checkmark  & $10000$ &&  \colorbox{pastelblue}{$31.48$} &	\colorbox{pastelgreen}{$41$}	& \colorbox{pastelgreen}{$33.82\%$}	 &&  \colorbox{pastelblue}{$31.80$} &	\colorbox{pastelblue}{$46$}	& \colorbox{pastelgreen}{$34.73\%$}  \\
$\times$  & $0$ && $30.70$ & $35$	& {$26.54\%$} && $30.66$  &	$39$ & \colorbox{pastelblue}{$31.18\%$}  \\
$\times$  & $10000$ && $30.86$ &	$34$ & \colorbox{pastelblue}{$26.64\%$} && $31.69$  & $46$	& $28.51\%$  \\
\bottomrule
\\
\end{tabular}
\caption{
\small\textbf{Conditional Evaluation: effect of mask normalization and $c_g$}: We fix the value of $M=2$ and ablate the effects of mask normalization and $c_g$. Note that we use $c_g=10000$ to report our results for temporal intrisnic denoising in the paper and $c_g=0$ corresponds to intrinsic denoising. Mask Normalization consistently leads to high CLIP-SIM score and coverage, and using mask normalization with $c_g=10000$ yields the best spatio-temporal control, as measured by mIoU.}
\label{tab:supp-cond-eval}
\end{table}

\begin{table}
\centering
\small
\begin{tabular}{cc|ccc}
\toprule
Mask Norm  & {$c_g$} & {{FID$\downarrow$}} & {{KID$\downarrow$}} & {\color{gray} {JeDi$^{\dagger}\downarrow$}} \\
& & & \\
\midrule
\checkmark  & $0$ & \colorbox{pastelblue}{$133.74$} &	\colorbox{pastelgreen}{$1.81\% \pm 0.06$} &  \colorbox{pastelgreen}{\color{gray} $1.60$}  \\
\checkmark  & $10000$ & \colorbox{pastelgreen}{$131.19$} & {$1.92\% \pm 0.06$} & {\color{gray} $1.86$} \\
$\times$  & $10000$ &  $144.85$ &  $2.88 \pm 0.09$ & {\color{gray} $1.77$} \\
$\times$ & $0$ & $135.49$ & \colorbox{pastelblue}{$1.86 \pm 0.06$} & \colorbox{pastelblue}{\color{gray} $1.61$} \\

\bottomrule 
\\
\end{tabular}
\caption{ \small \textbf{Unconditional Evaluation: effect of mask normalization and $c_g$}: We fix the value of $M=2$ and ablate the effects of mask normalization and $c_g$. Note that we use $c_g=10000$ to report our results for temporal intrisnic denoising in the paper and $c_g=0$ corresponds to intrinsic denoising. As explained in the main paper, the low JeDi scores for $c_g=0$ are misleading.}
\label{tab:supp-uncond-eval}
\end{table}

\subsection{Effect of gradient normalization}

\begin{figure}
\centering
\includegraphics[width=\textwidth]{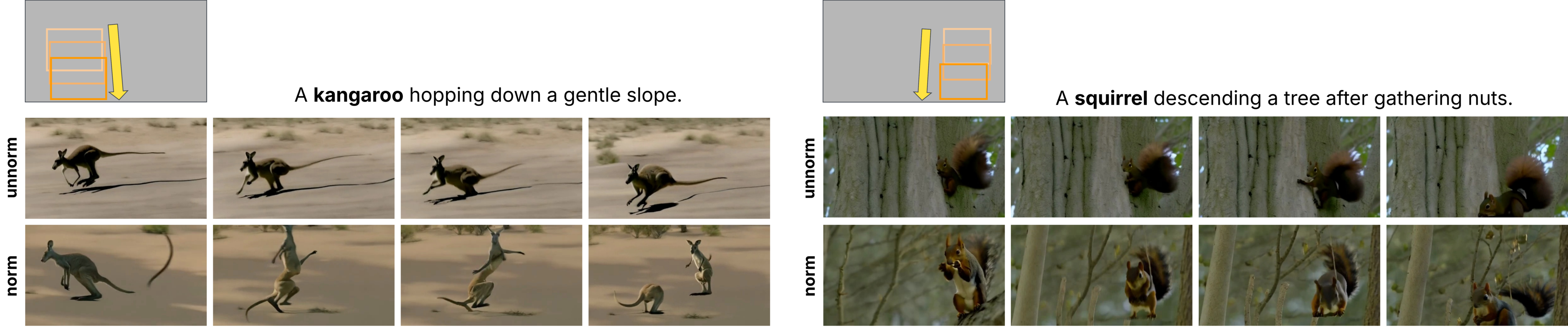}
\caption{\small \label{fig:cg-grad-norm} \textbf{Effect of gradient normalization:} We compare two variants of temporal intrinsic denoising: with gradient normalization (``norm'', bottom row) and without normalization (``unnorm'', top row). While ``norm'' yields improved results in some cases (right panel), it frequently introduces spatio-temporal distortions (left panel), whereas ``unnorm'' produces more stable and consistent outputs.} 
\end{figure}

In the current formulation of temporal intrinsic denoising, gradients are applied without explicit normalization. To study the effect of gradient scaling, we experiment with normalizing the gradients to have unit L2 norm before applying a fixed scale. For the normalized variant, we use 8 frozen steps and set the guidance strength to $c_g = 0.2$. However, we observe that normalizing the gradients before scaling often introduces spatio-temporal inconsistencies, as illustrated in \cref{fig:cg-grad-norm}. While this approach can occasionally improve spatiotemporal control (right panel), it frequently disrupts global coherence across frames (left panel). We posit that this effect arises because normalization preserves only the relative direction of the gradient while discarding information encoded in its absolute magnitude, which may be crucial for maintaining spatio-temporal coherence.

\subsection{Failure Cases}

\begin{figure}
\centering
\includegraphics[width=\textwidth]{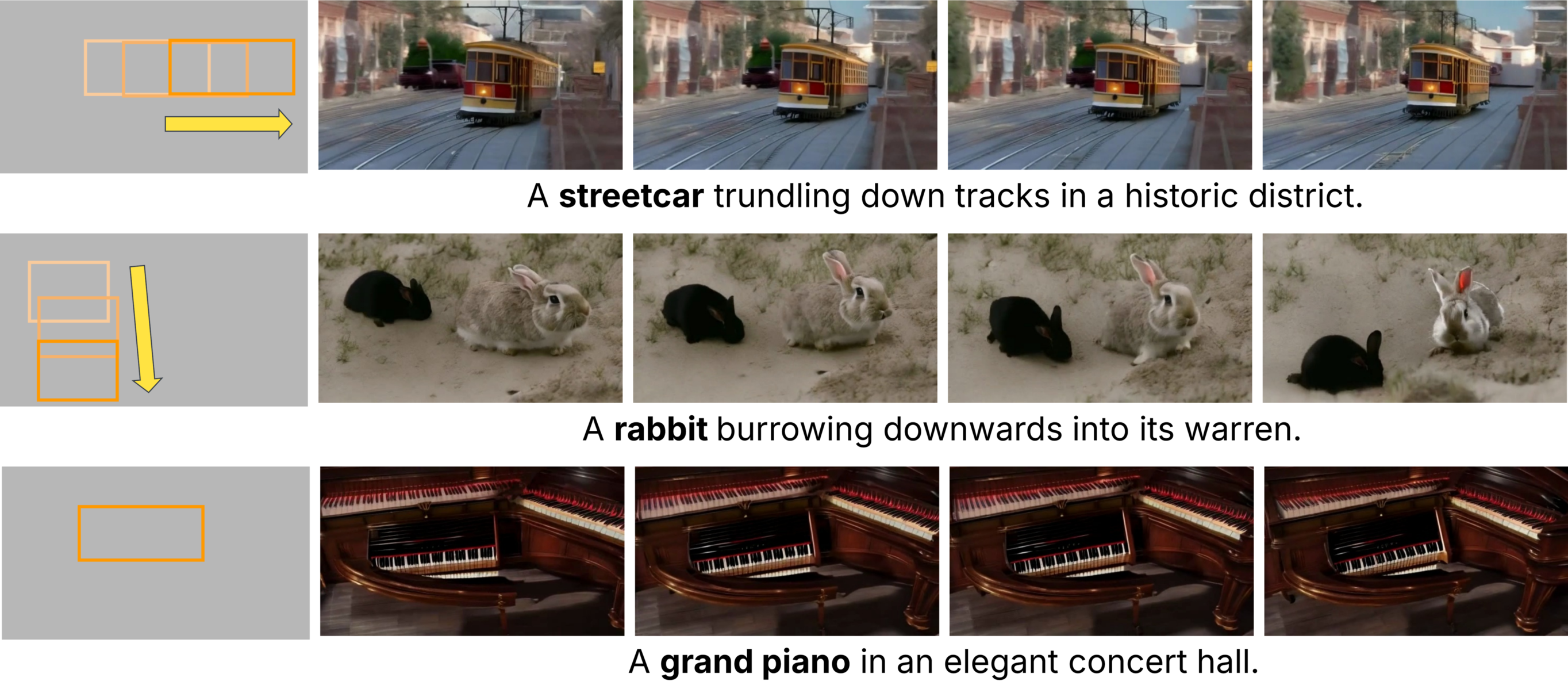}
\caption{\small \label{fig:failure_cases} \textbf{Failure Cases:} Top: The model fails to follow the user-guided trajectory. Middle: The model generates extra subjects in addition to the correct one following the trajectory. Bottom: The background appears unrealistic or inconsistent.
} 
\end{figure}

While our method outperforms prior work across several benchmarks, it is not without limitations. As shown in Figure~\ref{fig:failure_cases}, failure modes include deviations from the target trajectory, hallucination of extra subjects, and generation of unrealistic backgrounds. These cases highlight that no method is currently flawless and suggest important directions for future investigation.

\end{document}